\documentclass[11pt,letterpaper]{article}
\usepackage[top=1in,bottom=1in,left=1in,right=1in]{geometry}
\usepackage[usenames,dvipsnames]{color}
\definecolor{myblue}{rgb}{0,0.1,0.6}
\definecolor{mygreen}{rgb}{0,0.3,0.1}
\usepackage[colorlinks=true,linkcolor=black,citecolor=mygreen,urlcolor=myblue]{hyperref}
\pdfoutput=1

\usepackage{times}
\usepackage{latexsym}
\usepackage{amsmath}
\usepackage{multirow}
\usepackage{url}
\usepackage{ifthen}
\usepackage[pdftex]{graphicx}

\usepackage{array}
\usepackage{natbib}
\usepackage{chngpage}
\usepackage{stmaryrd}
\usepackage{amssymb}
\usepackage{amsmath}
\usepackage{graphicx}
\usepackage{lscape}
\usepackage{subfig}

\usepackage{natbib}
\usepackage{natbibspacing}
\bibpunct{(}{)}{;}{a}{,}{,}
\usepackage{sidecap}

\newenvironment{itemizesquish}{\begin{list}{\labelitemi}{\setlength{\itemsep}{-0.07em}\setlength{\labelwidth}{0.8em}\setlength{\leftmargin}{\labelwidth}\addtolength{\leftmargin}{\labelsep}}}{\end{list}}

\newcommand{\ensuretext}[1]{#1}
\newcommand{\arkcomment}[3]{}

\newcommand{\bo}{\ensuretext{\textcolor{blue}{\ensuremath{^{\textsc{B}}_{\textsc{O}}}}}}

\newcommand{\bocomment}[1]{\arkcomment{\bo}{#1}{blue}}

\title{
Learning Frames from Text with an Unsupervised Latent Variable Model
}

\author{
  Brendan O'Connor {\small(\url{http://brenocon.com})} \\
Machine Learning Department,
Carnegie Mellon University
}

\begin{document}
\maketitle

\begin{abstract}
\noindent
We develop a probabilistic latent-variable model to discover semantic
frames---types of events and their participants---from corpora.
We present a Dirichlet-multinomial model in which frames are latent
categories that explain the linking of verb-subject-object triples,
given document-level sparsity.
We analyze what the model learns, and compare it to FrameNet,
noting it learns some novel and interesting frames.
This document also contains
a discussion of inference issues,
including concentration parameter learning;
and a small-scale error analysis of syntactic parsing accuracy.
\end{abstract}

\noindent\emph{Note:} this work was originally posted online October 2012
as part of CMU MLD's Data Analysis Project requirement.
This version has no new experiments or results,
but has added some discussion of
new related work.

\section{Introduction}

Semantic frames---types of events and their participants---are a
key development in linguistic theories of semantics \citep{fillmore82} and, 
sometimes called ``scripts'' or ``schemata,''
have figured heavily in natural language understanding research \citep{schank_scripts_1977,lehnert_strategies_1982}.
There has been a recent surge  in interest in finding frames in text data, including
frame-semantic parsing \citep{das2010probabilistic,gildea2002automatic}
following the conventions of FrameNet \citep{fillmore2001frame},
and discovery of narrative structure \citep{chambers2009unsupervised}.
% also related is
% semi-supervised learning of relational knowledge \citep{carlson_toward_2010,mintz2009distant}.

In this paper, we seek to discover semantic frames---types of events or relations and their participants---from corpora, using  probabilistic latent-variable models.  
This approach focuses on verbs with their subjects and objects and is inspired by models of selectional preferences and argument structure.  Building on the framework of topic models \citep{blei_latent_2003},
we further leverage document context, exploiting an assumption that relatively few frames are expected to be present in a single document.  Our key contributions are in a new model in which
(1) frames are latent categories that explain the linking of verb-subject-object triples in a given document context; 
and (2) cross-cutting semantic word classes are learned, shared across frames.

Because there are many ways to define frames, we believe a data-driven approach that does not require human annotation is attractive, especially when considering a new domain of text,
or exploratory data analysis of a corpus.
We explore models built on a range of datasets, highlighting differences in what kinds of frames are discovered.

We also seek to evaluate what is learned by comparing to existing lexical resources,
introducing a novel evaluation methodology that compares to FrameNet, 
interpreting the model posterior as an alternative lexicon.   

% * Connect frame learning to topic models and rich corpus analysis
% speculative goal: structured opinion analysis \nascomment{seems too far a stretch at this point --- I would not play this up; instead I would play up the claim that there is an intellectual connection between frames/scripts and topic models}.  
 % For example, gender or race relations, e.g.~how have attitudes towards women changed through history?
 % not a binary sentiment polarity thing,
 % but rather, assumptions about the role of women in society
 % may be expressed in text through actions that women engage in or are involved in.
 % Need a latent argument structure model to discover these regularities.
 % this work is a first step in that direction

% \nascomment{penultimate paragraph of intro, which you will compress/consolidate into the conclusion as well:  key contributions of the paper.  I think these are (i) new unsupervised models for frame discovery, (ii) experiments on several large corpora, (iii) comparison to existing hand-built lexicon using a new methodology}

We begin by describing our models and relating them to models in the literature (\S\ref{se:models}).  
We discuss inference in \S\ref{se:inference} and experiments in \S\ref{se:experiments},
concluding with example results (\S\ref{se:examples}) and FrameNet comparison (\S\ref{se:fn}).

\section{Models} \label{se:models}

Verbs, subjects, and objects constitute a basic syntactic encoding
of actions, events, and their participants. \bocomment{frawley?}
We are interested in a modeling a dataset of document-VSO tuples,

\[ (\text{DocID}, w^{(verb)}, w^{(subj)}, w^{(obj)}) \]
% \bocomment{need to explain why.  close approximation of ARG0 and ARG1,
%   the most fundamental semantic roles.
%   Chambers and Jurafsky use them to build narrative schemas.
%   There are many frame theories (cite cite cite),
%   but verb-subject-object
%   is where to start for
%   inducing frames that are
%   coherent interrelated sets 
%   of
%   actions, actors, and participants.
% } \nascomment{+1 --- maybe say first what we want to model, then say that this boils down to a four-tuple representation of the data.  richer representations could come in later models, but for now this is what our models can see.}

We present two models to capture document and syntactic contextual information
in the generation of text.

\subsection{``Model 0'': Independent tuples}

Previous work in model-based syntactic distributional clustering,
usually aimed at modeling selection preferences,
has modeled syntactic tuples as independent
(or rather, conditionally independent given the model parameters).
\cite{pereira_distributional_1993} and \cite{rooth99}
model a corpus of (verb, object) pairs with a latent variable for each tuple,
and different word distributions for for each argument and class.
(Rooth et al.~experiment with different syntactic relations, but always use pairs; e.g.~(verb, subject).)

To situate our model, we slightly generalize these approaches to (verb, subject, object) triples, 
and add symmetric Dirichlet priors, as follows.
$\phi^{(arg)}_f$ denotes a word multinomial for argument type $arg$ and frame $f$;
and there are three argument types (verb, subject, object), treated completely separately.
\begin{itemizesquish}
\item
\textbf{Frame lexicon:}
Dirichlet prior $\beta$. For each $f=1..F$, sample three word multinomials:
$\phi^{(v)}_f, \phi^{(s)}_f, \phi^{(o)}_f \sim Dir(\beta)$
%, where e.g.~$\phi^{(v)}_f$ denotes the distribution for verbs under class $f$; and
\item
\textbf{Tuple data:}
For each tuple $i=1..N$, 
  \begin{itemizesquish}
  \item Draw its frame indicator $f_i$ (from a fixed prior),
  \item Draw the three words from their repsective multinomials:
  $w^{(v)}_i \sim \phi^{(v)}_{f_i};\ w^{(s)}_i \sim \phi^{(s)}_{f_i};\ w^{(o)}_i \sim \phi^{(s)}_{f_i}$
  \end{itemizesquish}
\end{itemizesquish}
This approach models every tuple independently, with no document or other context;
the only thing shared across tuples are the per-class argument word distributions.
More recent work extends the Rooth approach to model other syntactic relation pairs
\citep{oshea10}
and web-extracted triple relations \citep{ritter10}.
These works make several variations to
the probabilistic directed graph, giving the verb a more central role;
we instead stick with Rooth's symmetric setup.

\subsection{Model 1: Document-tuples}
We would like to use document context to constrain the selection of frames.
Following the intuition of latent Dirichlet allocation \citep{blei_latent_2003}---that 
each particular document
tends to use a small subset of available latent semantic factors (``topics'')---we propose 
that 
a document's frames
are similarly drawn from a sparsity-inducing Dirichlet prior.
Our document-tuple model uses the same frame lexicon setup as above, but enriches the document generation:

\begin{itemizesquish}
\item $F$ frames, and Dirichlet priors $\alpha,\beta$
\item \textbf{Frame lexicon:}
  For each frame $f \in 1..F$, and argument position $a\in\{1,2,3\}$,
    \begin{itemizesquish}\item
    Draw word multinomial $\phi^{(a)}_f \sim Dir(\beta)$
    \end{itemizesquish}
  \item \textbf{Document-tuple data:} For each document $d \in 1..D$,
  \begin{itemizesquish}
  \item Draw frame multinomial $\theta_d \sim Dir(\alpha)$
  \item For each tuple $i$ in the document,
    \begin{itemizesquish}
    \item Draw frame indicator $f_i \sim \theta_d$
    \item Draw word triple: for each argument position $a \in \{1,2,3\}$,
      \begin{itemizesquish}
      \item Draw $w^{(a)}_i \sim \phi^{(a)}_{f_i}$
      \end{itemizesquish}
    \end{itemizesquish}
  \end{itemizesquish}
\end{itemizesquish}

Note that in the limiting case as $\alpha \rightarrow \infty$, Model 1 collapses into 
the independent tuple model, where document context is irrelevant.
In our experiments, we fit $\alpha$ with posterior inference, and it prefers to have relatively 
low values, giving orders of magnitude better likelihood than a high $\alpha$---implying that the document-level
sparsity assumption better explains the data than an independent tuple hypothesis.

From one perspective, LDA's ``topics'' have been renamed ``frames.''
Computationally, this is a minor difference,
but is significant from an NLP perspective,
% but from an NLP persepctive, is very important,
% but is a big difference linguistically,
since we are asking the latent variable to do something else than LDA has it do---it
now models syntactic argument selection as well as document-level effects.

The document-level mixing is potentially useful for applications,
because it gives a hook into a vast literature of topic models
that jointly model text with many types of document-level metadata
such as time, space, arbitrary metadata, etc.~(e.g.~\cite{blei2006dynamic,eisenstein_latent_2010,mimno2008topic})
In this respect, this model could be seen as a ``semantic'' or ``syntactic tuple''
topic model, along the lines of previous work
that has added various forms of sentence-internal
structure to LDA's document generation process,
such as
bigrams~\citep{wallach2006ngrams},
HMM's~\citep{griffiths2005integrating},
or certain forms of syntax~\citep{boydgraber2008}.

%   our twist: LDA is more like frame semantics than it is super-local syntactic contexts (wallach, steyvers),
%   and frame semantics is better represented through dependencies than constituents (boyd-graber)

% Typically, models in natural language processing ignore the document ID,
% but it easy to incorporate in this approach.
% oconnor2010mixture}
% These models nearly always use bag-of-words assumptions;
% frame models can be plugged into them.
% \bocomment{this document-level mixing gives a hook into the wide world of topic models
%   that jointly model text with document-level metadata:
%   time, space, authors, etc.
% } \nascomment{give a few cites but this will be familiar}

\begin{figure}\centering
\includegraphics[width=2.7in]{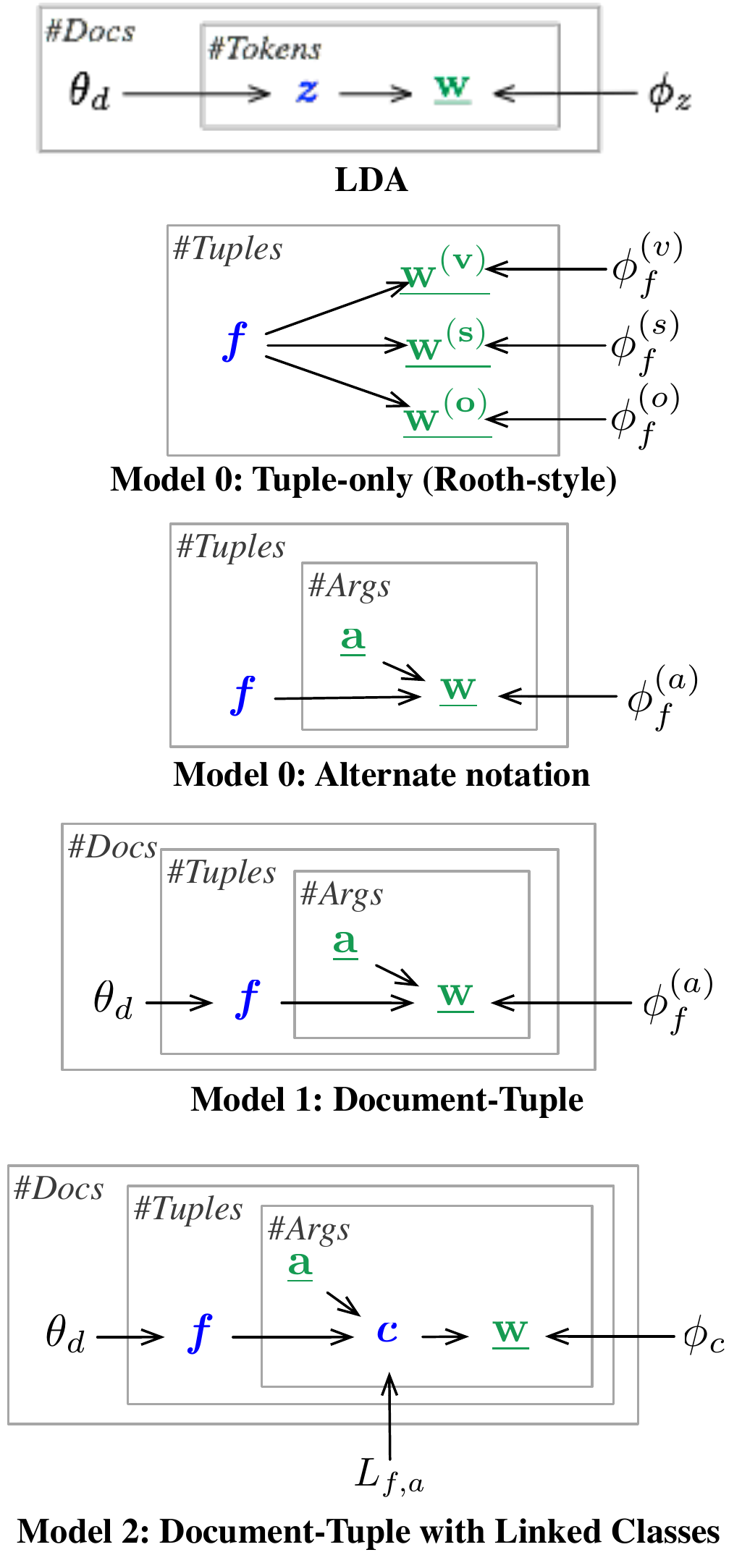}
\caption{Probabilistic directed graphs for prior work and our models.
Dirichlet priors are omitted for brevity.
\textbf{\textcolor{blue}{Blue}} variables are latent,
and resampled through collapsed Gibbs sampling;
\textbf{\textcolor{ForestGreen}{\underline{green}}} variables are observed.
}
\label{diagram_gm}
\end{figure}
\begin{figure}\centering
\includegraphics[width=1.1in]{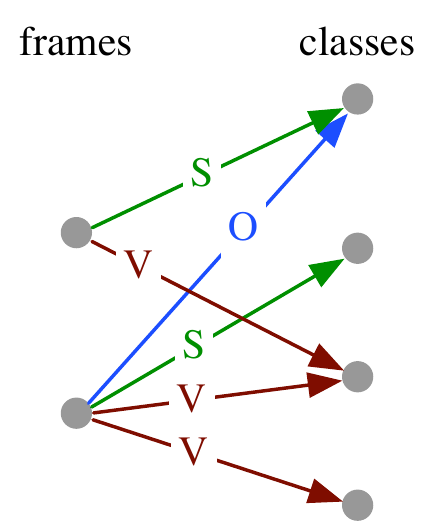}
\caption{Example fragment of the sparse linking array $L_{f,a,c}$ as a labeled bipartite graph.
For example, the \emph{S}-labeled edge connecting frame 1 and class 1
indicates $L_{1,2,1} > 0$ (assuming subject corresponds to $a=2$).
}
% Sparsity in the graph is due to Dirichlet prior on $L_{f,a}$ distributions. \nascomment{shouldn't each frame have some outbound arcs?}} 
\label{linker}
\end{figure}

\subsection{Model 2: Cross-cutting semantic classes}

The frames in Model 1 share no word statistics with one another,
so each has to relearn lexical classes for its arguments.
We address this by introducing a latent word class variables $c$
for every token.
Every frame has preferences for different classes at different argument positions,
so word classes can be shared across frames.
% The hope is that such classes
% may constitute higher level ``roles'' for a script-like high-level frame.
% the model's ``frames'' are really microframes, so share across them.
% for clarity we always refer to these things as ``classes''.

The Model 2 generative process is:
\begin{itemizesquish}
\item $C$ word classes, $F$ frames, and Dirichlet priors $\alpha,\beta,\gamma_1,\gamma_2,\gamma_3$
\item \textbf{Frame lexicon:}
\begin{itemizesquish}
  \item For each class $c\in1..C$, 
    \begin{itemizesquish}\item
    Draw word multinomial $\phi_c \sim Dir(\beta)$
    \end{itemizesquish}
  \item For each frame $f\in 1..F$, and argument position $a \in \{1,2,3\}$,
    \begin{itemizesquish}\item
    Draw the ``linker''  $L_{f,a} \sim Dir(\gamma)$, a multinomial over word classes: $L_{f,a} \in Simplex(C)$.
    \end{itemizesquish}
\end{itemizesquish}
\item \textbf{Document-tuple data:}
\begin{itemizesquish}
  \item For each document $d \in 1..D$,
    draw frame multinomial $\theta_d \sim Dir(\alpha)$
  \item For each tuple $i$ for document $d$,
  \begin{itemizesquish}
    \item Draw frame indicator $f_i \sim \theta_d$
    \item Draw word triple: for each argument $a \in \{1,2,3\}$,
    \begin{itemizesquish}
      \item (if $a$ is null in this tuple, skip)
      \item Draw class $c^{(a)} \sim L_{f_i,a}$
      \item Draw word $w^{(a)} \sim \phi_{c^{(a)}}$
    \end{itemizesquish}
  \end{itemizesquish}
\end{itemizesquish}
\end{itemizesquish}

Central to this model is the ``linker'' $L_{f,a,c}$, a multidimensional array 
of dimensions $(F,3,C)$, 
\bocomment{took out ``tensor'', physics ppl think it's too grandiose a name for a mere multidim array}
which says which word classes are likely for a given frame-argument combination.
We show a schematic diagram in Figure~\ref{linker}, where edges represent high 
(or just significantly nonzero) probabilities.
A word class that is a subject for one frame may be an object for another.
A frame may have multiple classes for an argument position,
though in practice the number is relatively small, due to the Dirichlet prior 
(and, as we will see,
it naturally turns out sparse when the prior is fit to the data).

Note that Model 1 can be seen as a version of Model 2 with a deterministic linker:
every $(f,a)$ pair is bound to one single, unique class; i.e.~a point mass at~$c=3(f-1)+a-1$.
Therefore Model 1 corresponds to a version of Model 2 with 
a deterministic linking tensor $L_{f,a,c}$, where every $(f,a)$ slice has only a single non-zero element.
This view also suggests a non-parametric version of Model 2 
corresponding to a stick-breaking prior for $L_{f,a}$;
this would be an
interesting extension, but we use a simple fixed Dirichlet prior
in this work.

\cite{titov_bayesian_2011}'s model of semantic frames
also uses cross-cutting word classes, embedded in a more complex model
that also learns clusters of syntactic relations,
and recursively generates dependency trees.
\cite{grenager_unsupervised_2006} and \cite{lang2010unsupervised}
present related models for unsupervised PropBank-style
semantic role labeling,
where a major focus is grouping or clustering syntactic 
argument patterns.
Other related models include \citep{Gildea2002Verbs,Titov2012SRL};
most relevantly, the recently published work of \cite{Modi2012Frames}
compares a similar Bayesian model directly to FrameNet,
and \cite{Cheung2013Frames} integrates discourse into a frame model,
capturing the script-style structures that were previously explored in ad-hoc approaches by \cite{Chambers2011Templates}.

Finally, while we do not enforce any relationship between syntactic argument position
and the classes, in practice, most classes are exclusively either verbs or nouns,
since words that can be verbs often cannot appear as nouns.

% \subsection{DELETE this, has been folded into throwaway comment above}
% \bocomment{ok can now kill this section\ldots}
% The relationship between Models 1 and 2
% can also be motivated by describing Model 1
% in terms of a Dirichlet process.
% \bocomment{this is weird\ldots should first note deterministic linker, THEN the DP story.
% but kinda irrelevant?}
% For Model 1, 
% let $\phi_{f,a}$ be the multinomial over words for frame $f$ and argument position $a$.
% If $\phi_{f,a} \sim DP( G_0, \gamma )$ 
% % (with a base distribution corresponding to the prior over word multinomials),
% then as $\gamma \rightarrow 0$, every $\phi_{f,a}$ draw will be unique, and concentrated 
% on one single word distribution.
% We can introduce indicator variables $c=1..3F$, as well as $3F$
% linking multinomials $L_{f,a}$
% each having point mass at $c(f,a)=3(f-1)+a-1$.
% Then $\phi_{f,a}$ can be relabeled as $\phi_{c(f,a)}$.
% % This is just a variant of the standard derivation 
% % of the DP CRP representation (e.g.~Neal 2000).
% Therefore Model 1 corresponds to a version of Model 2 with 
% a deterministic linking tensor $L_{f,a,c}$, where every $(f,a)$ slice has only a single non-zero element.
% 
% This view also suggests a non-parametric version of Model 2 
% corresponding to a stick-breaking prior for $L_{f,a}$;
% this would be an
% interesting extension, but we use a simple fixed Dirichlet prior
% in this work.

\section{Inference} \label{se:inference}

Through Dirichlet-multinomial conjugacy, we can use collapsed Gibbs sampling for inference
% which is a well-established for these types of grouped Dirichlet-multinomial mixture models
\citep{griffiths_finding_2004,neal_bayesian_1992}.
The Gibbs sampling equations are instructive. For Model 1, it is:

\newcommand{\factorpad}{\vspace{0.05in}}
\factorpad\begin{center}
\includegraphics[height=0.6in]{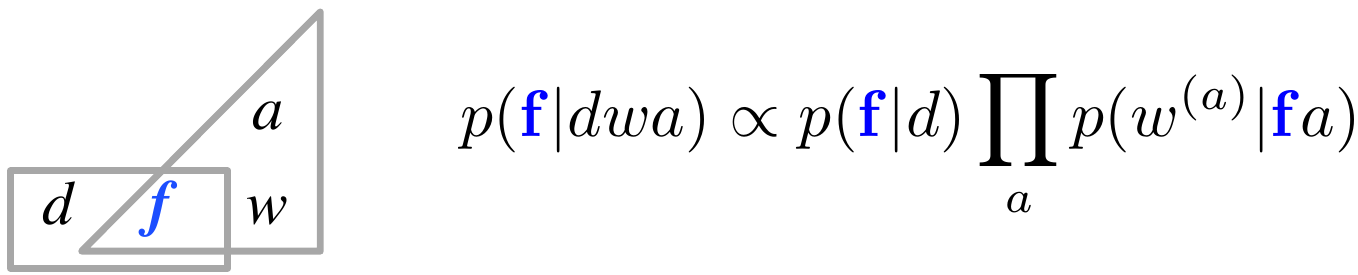}
% \factorpad
\end{center}

\noindent This diagram of CPT factors shows the two
soft constraints the Gibbs sampler works to satisfy:
the left term $p(f|d)$ tries to ensure document-frame coherency---it exerts pressure
to select a frame used elsewhere in the document.
The second term $p(w|fa)$ exerts pressure for syntactic coherency---to
choose a frame that has compatibility with all the syntactic arguments.
Thus Model 1 combines selectional preferences with document modeling.

Model 2's Gibbs sampling equations are

\factorpad
\begin{center}
\includegraphics[height=0.6in]{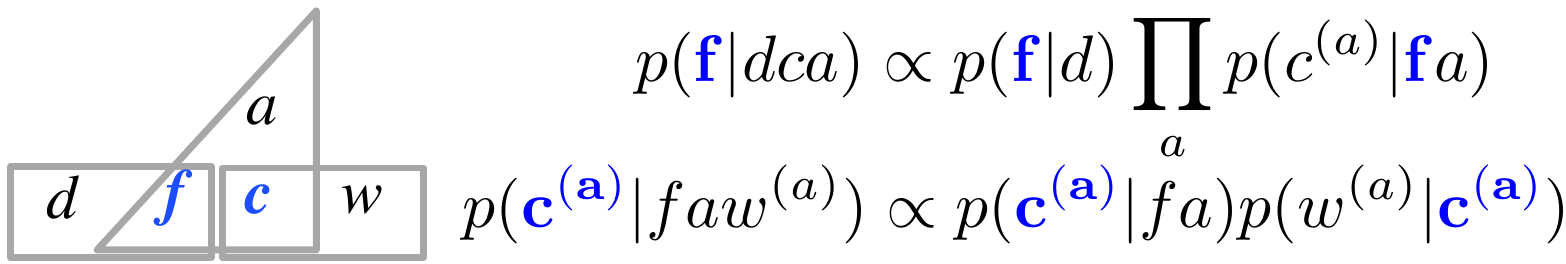}
% \factorpad
\end{center}

\noindent In these factor diagrams, the boxes correspond to
the maximal dimensional count tables the sampler has to maintain;
for example, the tables $C(d,f)$ and $C(f,a,w)$ for Model 1.  They are only the maximal, not all, count tables, since some rollups also have to be maintained for CGS denominators; e.g.~$C(d)$ and $C(f,a)$ here.

We resample the symmetric Dirichlet priors $\alpha,\beta,\gamma_1,\gamma_2,\gamma_3$
with slice sampling 
\citep{neal2003slice}
every 100 iterations, 
using a vague hyperprior.
We found the choice of hyperprior
  (either vague gamma or improper uniform) made little difference.

\begin{figure}
  \begin{center}
  \includegraphics[width=5in]{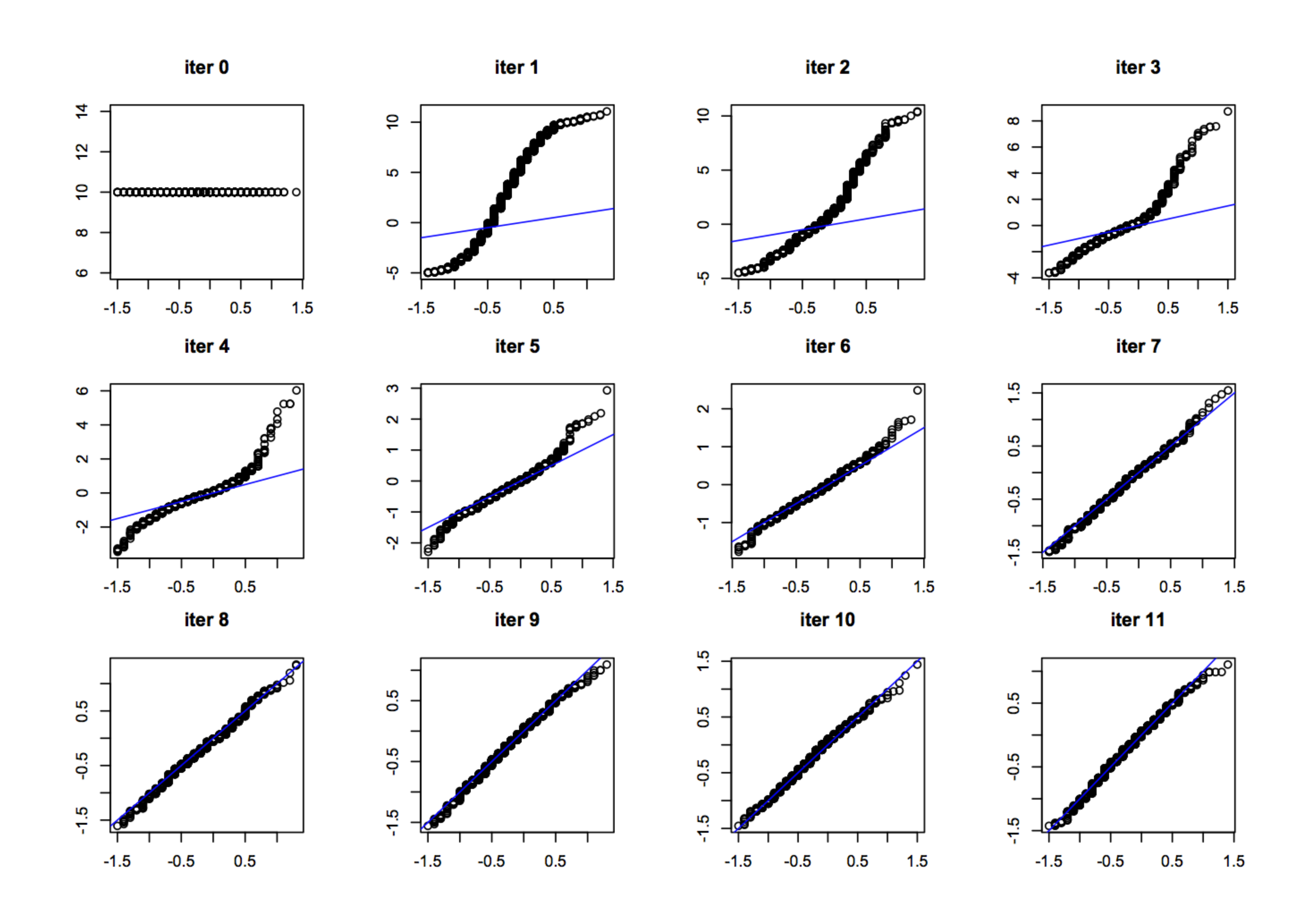}
  \end{center}
  \caption{QQ-plots of 500 slice sampling chains, showing their
  distribution converging to the true posterior.
  Each plot is the QQ-plot of 500 different chain states, all at
  the same iteration, against the exhaustively calculated posterior.}
  \label{fig:qqplots}
\end{figure}

The slice sampler was run for 10 iterations
for each parameter, a number chosen based on a simulation experiment:
from a fixed $\alpha$,
we generated a Dirichlet-multinomial dataset (10 groups, 10 observations each),
then ran 500 independent slice sampling chains for the variable $\log(\alpha)$,
assessing the cross-chain distribution
of states at a single timestep against the true posterior
(the latter computed with near-exact grid approximation on the DM likelihood)
via QQ-plots shown in Figure \ref{fig:qqplots}.
MCMC theory says that after burn-in you can stop the chain to have a single
independent draw from the posterior, which implies these QQ-plots will become
flat; it appears 7 iterations was enough in this experiment.
We chose here a very bad initializer $\alpha=e^{10}$,
compared to the true MAP $e^{0.9}$; a good initialization at $\alpha=e^1$ had
posterior convergence after just 2 iterations.
(This analysis technique was inspired by \cite{cook06}'s Bayesian software validation method.)

The hyperparameter sampling substantially improves likelihood.
This is in line with \cite{Wallach2009Asymmetric} and \cite{Asuncion2009},
which found that
learning these Dirichlet hyperparameters for LDA gave much better
solutions than fixing them---this makes sense, since a researcher can't
have much idea of good values to arbitrarily pick.  
(An alternative approach,
cross-validated grid search (used in e.g.~\citep{Griffiths2004CGS}),
is far more expensive than fitting the
hyperparameters on the training data---it is infeasible once there are several
concentration paramaters, as there are here.
\citeauthor{Asuncion2009} found that direct hyperparameter learning
performed as well as grid search.)
Furthermore, the recent work of
\cite{Chuang2013Diagnostics} found that hyperparameter learning also gave
near-optimal \emph{semantic} coherency for LDA (according to expert judgments
of topical semantic quality).
These works use a slightly different method for hyperparameter learning
(empirical Bayes optimization, as opposed to MCMC sampling); 
but we suspect the
choice of method makes little difference.  What matters is getting away from
the human-chosen initial value, which will be pretty poor.

\begin{figure}
  \begin{center}
    \includegraphics[width=2.8in]{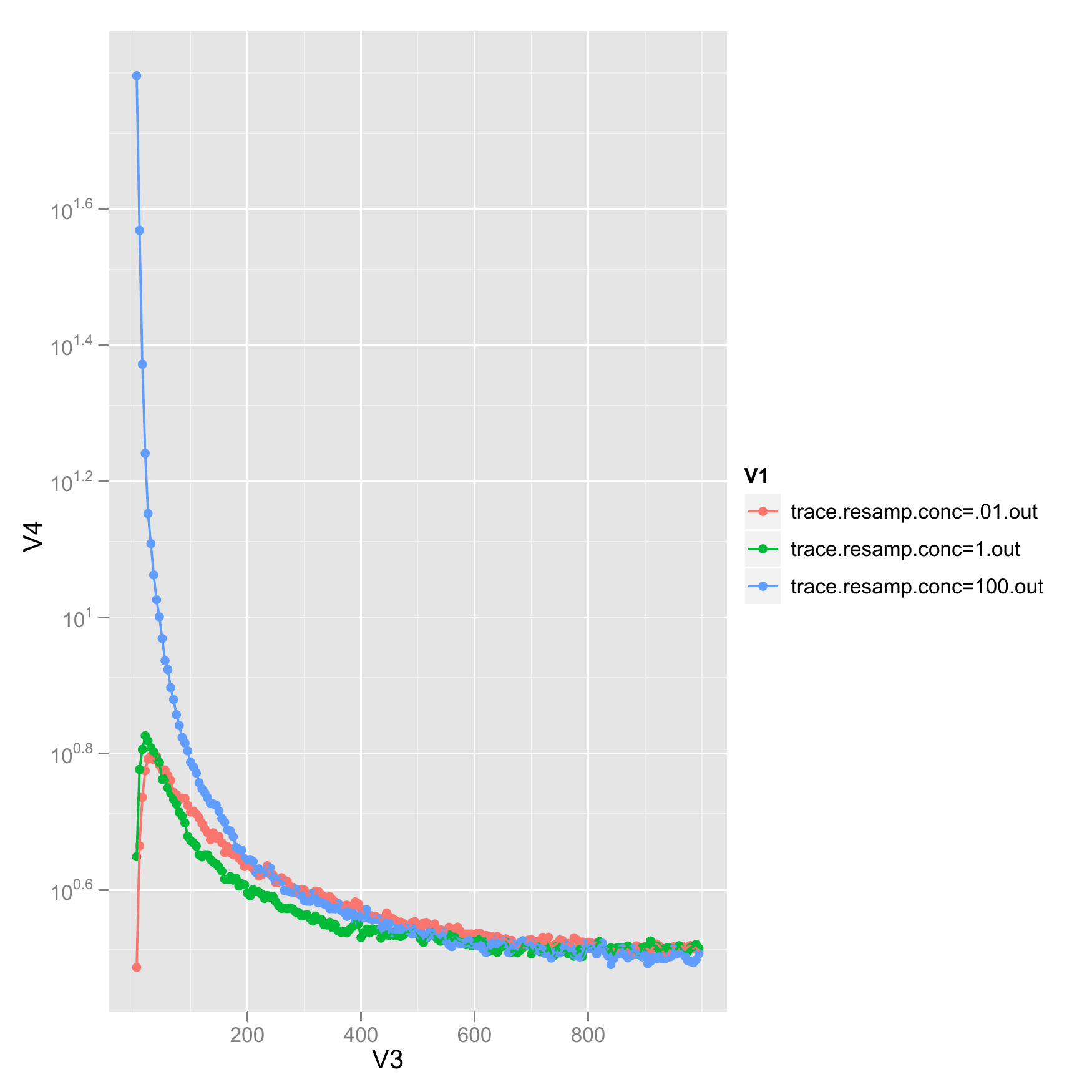}
    \includegraphics[width=3.2in]{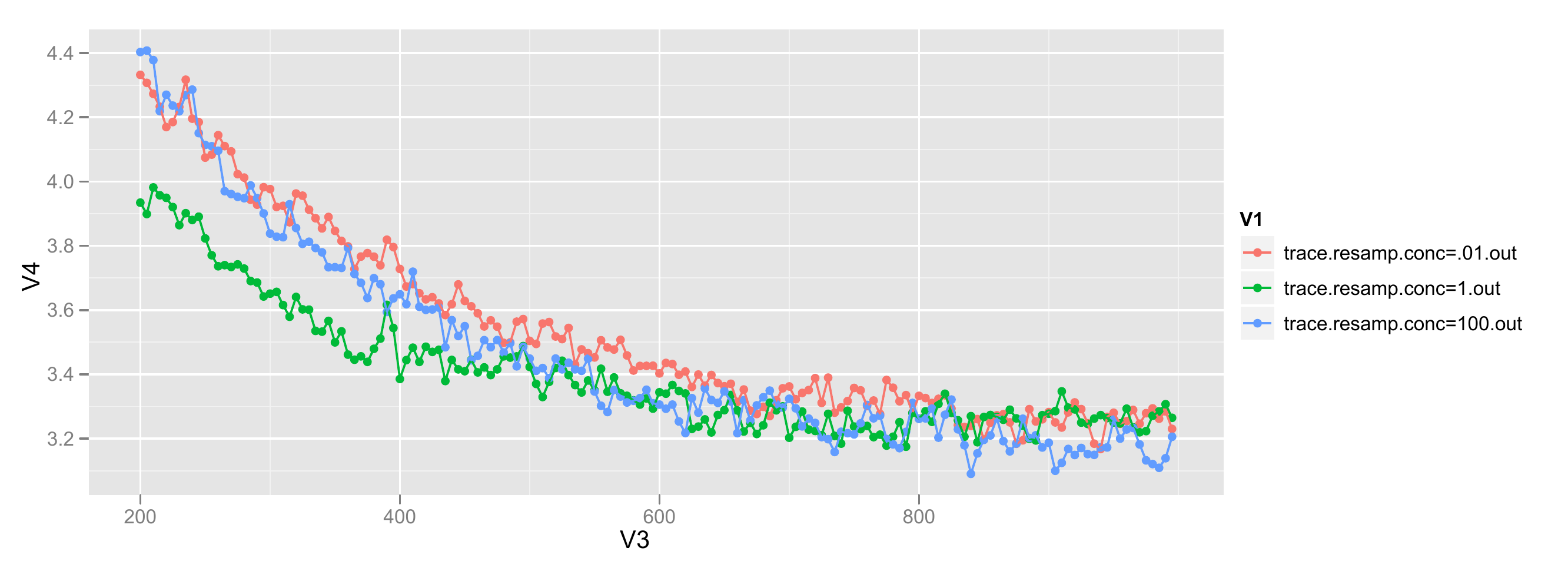}
  \end{center}
  \caption{$\alpha$ parameter values over time, when being resampled, from
  three very different initial positions $\alpha=0.01,\ 1,\ 100$.  The left
  plot shows the entire history; the right shows from iteration 200 until the
  end.  
    This is Model 1 on a subset of the \textsc{CrimeNYT} corpus.
  }
  \label{fig:dir_resamp}
\end{figure}

Interestingly, most of the movement tends to happen early in the MCMC chain,
then the hyperparameter stabilizes as the rest of the model is still moving.
For one experimental setting (a subset of \textsc{CrimeNYT}, described below),
we checked if the outcome was initializer dependent by starting three different
MCMC chains that were identical except for three different $\alpha$ initializers:
Figure \ref{fig:dir_resamp}.
Reassuringly, they all converged on the same region of values. This robustness
to initialization was exactly what we wanted.

For larger datasets, we implemented a parallelized sampling scheme
for $f$ and $c$
similar to
\cite{newman2009distributed}
where individual processors use stale counts and synchronize once per iteration
by sending count update messages to all other processors.\footnote{
We use from 12 to 64 CPU cores in some cases. Implementation is in Python/C/MPI.
We use MPI's \emph{AllReduce} operation for count table synchronization,
inspired by
\cite{Agarwal2011Allreduce}, which notes
it is a very convenient approach
to parallelize an initially
single-processor implementation of a machine learning algorithm.
}

In experiments, we ran the Gibbs sampler for at least 5,000 iterations, and up to 20,000 as time permitted.

\section{Experiments} \label{se:experiments}

\subsection{Datasets}
We use datasets from the New York Times, Wall Street Journal, and the Brown corpus for experiments.
The New York Times Annotated Corpus~\citep{nyt_ldc}
contains articles from 1987 to 2007 that have been extensively categorized by hand,
each having multiple labels.
We use it to extract two subcorpora.
First,
inspired by examples in the crime reporting domain
of extracting narrative event structures
\citep{chambers2009unsupervised}
and work on the FrameNet corpus \citep{fillmore2001frame},
we select news articles for which any category labels 
contain any of the words \emph{crime}, \emph{crimes}, or \emph{criminal},
resulting in a targeted corpus of articles (\textsc{CrimeNYT}).
We wanted to see if the model can learn specific types of actions and noun classes for that domain.
Second, we take a uniformly drawn subsample of all articles (\textsc{UnifNYT}).

The NYT article texts are processed with the Stanford CoreNLP
software\footnote{
  \url{http://nlp.stanford.edu/software/corenlp.shtml}
  % \url{http://nlp.stanford.edu/software/}
}
using default settings;
we use its
sentence segmentation, tokenization, part-of-speech tagging, lemmatization, parsing, and dependency conversion
(the \emph{CCprocessed} variant).

Finally, we also performed experiments on two pre-parsed corpora from the Penn Treebank
\citep{marcus_building_1994},
cointaining tokenizations, part-of-speech tags, and parses:
The Wall Street Journal (all sections; \textsc{WsjPTB}),
and the PTB's subset of the Brown corpus---consisting mainly of literature and essays (\textsc{BrownPTB}).  
We used the Stanford software to
convert PTB constituent structures to dependencies and produce lemmatizations.
%\footnote{Their
  % lemmatizer is a port of \emph{morpha} CITE: URL}
These are substantially
smaller than the NYT datasets; see Table~\ref{table:datasets}.

\begin{table}
  \centering
  \begin{tabular}{|l|r|r|r|r|}
    \hline
    Corpus            & \#Docs & \#Sent & \#Tok & \#VSO \\
    \hline
    \textsc{CrimeNYT} & 27k    & 789k   & 20M   & 1.3M \\
    \textsc{UnifNYT}  & 60k    & 2.4M   & 14M   & 5.3M \\
  % 60,302
% brenocon@tg-login1:~/sem/semdoc/data/nyt/unif_sample.crime1 % pv downsample=0.028.D=60k.semtuple | cut -f3 | wc
% 2,416,058 14,496,348 134728784

% brenocon@tg-login1:~/sem/semdoc/data/nyt/unif_sample.crime1 % pv full.downsample=0.14.D=245k.semtuple | cut -f3 | wc 
% 8,773,715 52642290 509289446
    \textsc{WsjPTB}   & 2,312  & 49k    & 1.2M  & 78k \\
    \textsc{BrownPTB} & 192    & 24k    & 459k  & 27k \\
    \hline
  \end{tabular}
  \caption{Datasets used in experiments.  
    The number of documents and VSO tuples are relevant to the model;
    the number of sentences and original tokens are not.
    The tuple count includes partial tuples (missing either subject or object).
  % $k=10^3,M=10^6$
  }
  \label{table:datasets}
\end{table}

From the dependencies, we extract active voice verb-subject-object tuples of form 
$(w^{(v)}, w^{(s)}, w^{(o)})$,
$(w^{(v)}, w^{(s)}, null)$, or
$(w^{(v)}, null, w^{(o)})$,
by finding instances of a verb word $w^{(v)}$ with POS tag beginning with VB,
and having a noun dependent with relation \emph{nsubj} or \emph{dobj}.
If both an \emph{nsubj} and \emph{dobj} child exist, take them both to form a full VSO triple; 
otherwise make a VS\_ or V\_O pair.
(Most are incomplete: in \textsc{CrimeNYT},
19\% of tuples are VSO, while 43\% are VS\_ and 38\% are V\_O.)
If multiple \emph{nsubj}'s exist, we arbitrarily take only one; similarly with \emph{dobj}.
(It may be better to take all possible tuples in this situation.)
% This can be intended behavior with Stanford Dependencies' coordination collapsing. 
% }

We performed error analysis of the parser and syntactic extraction,
by selecting a random sample of extracted V-S-O tuples from the \textsc{CrimeNYT} corpus
to manually assess for accuracy.  A subset are shown in Table \ref{parsing_accuracy}.  
We annotated 40 tuples, in context in their respective sentences,
consulting the Stanford Dependency papers 
\citep{de_marneffe_generating_2006,de_marneffe_stanford_2008},
which have
clear linguistic definitions of the grammatical relations, their conventions for analyzing compound verbs, etc.
Out of 40 tuples, we found 30 had the subject and/or object arguments correct; 6 had one or both wrong; and 4 were incomplete (missing either subject or object)---$75\%$ precision.

\begin{table}
  \centering
  \begin{tabular}{|m{1.8in}|m{3.2in}|}
    \hline
    correct? & text and VSO tuple \\
    \hline
    \textsc{right}: for ``workers,'' only use single head word of the noun phrase
    & In less than an hour , the police and rescue unit ( workers )$_{subj}$ [ found ]$_{verb}$ the ( organ )$_{obj}$ in the tall grass of the field , packed it in ice and took it to the hospital . 
    \\ \hline
    \textsc{right} &
    Mrs. ( Bissell )$_{subj}$ , she said , never [ learned ]$_{verb}$ what her husband did with the money . 
    \\ \hline
    \textsc{incomplete}: ``he'' should be subject of ``defrauded'' since SD uses content verb as head
    & Asked why he had [ defrauded ]$_{verb}$ the insurance ( company )$_{obj}$ and been partners with the mob \ldots
    \\ \hline \textsc{wrong}: lists are a known problem case for current
    parsers
    &
    Guilty on Five Charges Mr. Garcia was found guilty on all five charges against him : ( theft )$_{subj}$ of heroin , possession with intent to [ distribute ]$_{verb}$ heroin , narcotics conspiracy and two ( counts )$_{obj}$ of money laundering . 
    \\
    \hline
  \end{tabular}
  \caption{Example extracted tuples and our annotations in their original sentences (in tokenized form).}
  \label{parsing_accuracy}
\end{table}

% \subsection{Training}
% 
% We ran the Gibbs sampler for at least 5,000 iterations, and up to 20,000 as time permitted.
% % Likelihood starts leveling off, though the subjective quality of the clusters
% % see
% 
% MCMC convergence for mixture models
% can be difficult to diagnose
% \citep{KassRoundtable}
% since the sampler may wander through different mixture components 
% even when it is reasonably mixing.
% 
% \bocomment{GRAPHS}
% graphs:
% loglikelihood and priors
% 
%  * for small dataset .. many many iterations
% 
%  * multichain alpha convergence -- on small dataset
% 
%  * for crimeNYT .. 5000 iterations

\section{Example Results} \label{se:examples}

% We fit Model 2 to all datasets and Model 1 to some of them.
Here we focus on inspecting the posterior results from Model 2 on
the \textsc{CrimeNYT} dataset, with 400 frames and classes.

Table \ref{legislation} shows one particular interesting frame, which we intepret as ``legislation.''
We show its argument linking patterns by listing, for each argument position,
the most common classes appearing there.
These linking probabilities ($L_{f,a,c}=p(c|fa)$)
are show in the left column---the linkers are quite sparse,
with most of the probability mass contained in the shown classes.

For each class, we show the most common words.
Interestingly, the frame distinguishes different types of actions and actors
where laws or procedures are the object.
In fact, it nearly describes a sequential ``script'' of actions 
of the process of creating, enforcing, and modifying laws---but
the model has no knowledge of anything sequential.
This happens simply because there are several sets of actors
that perform actions upon laws,
and the model can picks up on this fact;
the model can see individual events in such a process,
but not the structure of the process itself.

\newcommand{\frametextwidth}{3.65in}
\newcommand{\frameinterpwidth}{1in}
\newcommand{\framestart}[2]{
  \vspace{0.1in}
  \begin{tabular}{|p{0.5in}|p{\frametextwidth}|p{\frameinterpwidth}|}
    \hline
    \multicolumn{3}{|c|}{$f$=#1: \emph{#2}} \\
    \hline
    $a,c,$ {\scriptsize $p(c|fa)$} & \emph{top words} & \emph{interp.} \\
    \hline
}
\newcommand{\syncolor}[2]{
  \ifthenelse{\equal{#1}{v}}{\textcolor{red}{#2}} 
      {
        \ifthenelse{\equal{#1}{s}}{\textcolor{ForestGreen}{#2}}
        {
          \ifthenelse{\equal{#1}{o}}{\textcolor{blue}{#2}}
          {}%
        }%
  }%
}
\newcommand{\R}[5]{ 
  \syncolor{#1}{(#1)} #2 {\scriptsize (#3)} & 
  \parbox{\frametextwidth}{\vspace{0.01in} 
    \syncolor{#1}{#5}
  \vspace{0.03in}} & 
  \parbox{\frameinterpwidth}{\vspace{0.01in} 
    #4 
  \vspace{0.01in}} 
  \\ 
  \hline
}
\newcommand{\frameend}[0]{ 
  \end{tabular} 
  \vspace{0.03in}
}

\newcommand{\fframetextwidth}{2.0in}
\newcommand{\fframeinterpwidth}{0.38in}
\newcommand{\fframestart}[2]{
  \vspace{0.1in}
  \begin{tabular}{|p{0.25in}|p{\fframetextwidth}|}
    \hline
    \multicolumn{2}{|c|}{$f$=#1: \emph{#2}} \\
    \hline
    $a,c,$ {\scriptsize $p(c|fa)$} & \emph{top words} \\ 
    \hline
}
\newcommand{\ffr}[4]{ 
  \syncolor{#1}{(#1)} #2 {\scriptsize (#3)} & 
  \parbox{\fframetextwidth}{\vspace{0.01in} 
    \syncolor{#1}{#4}
  \vspace{0.03in}} 
  \\ 
  \hline
}
\newcommand{\fframeend}[0]{ 
  \end{tabular} 
  \vspace{0.03in}
}
  % & 
  % \parbox{\frameinterpwidth}{\vspace{0.01in} 
  %   #4 
  % \vspace{0.01in}} 

\begin{table}
\centering
  \footnotesize
  % In [17]: import m2

% In [16]: m=m2.load_last("exper2/trace.crime2.KR=400.last")

% In [10]: C=m.C_frame_arg_role[286,0]
% 
% In [11]: C[(-C).argsort()[:5]] / C.sum()
% Out[11]: array([ 0.45234281,  0.20064077,  0.11393672,  0.04285142,  0.04144974])
% 
% In [12]: C=m.C_frame_arg_role[286,1]
% 
% In [13]: C[(-C).argsort()[:2]] / C.sum()
% Out[13]: array([ 0.59810501,  0.14488749])
% 
% In [14]: C=m.C_frame_arg_role[286,2]
% 
% In [15]: C[(-C).argsort()[:5]] / C.sum()
% Out[15]: array([ 0.80376085,  0.08751205,  0.05834137,  0.02844744,  0.00626808])

\framestart{286}{Legislation}
\R{v}{242}{0.45}{passage/enactment}{ 
pass have enact impose adopt extend eliminate increase toughen abolish amend need use establish change fail strike consider restore ease}
\R{v}{34}{0.21}{enforcement and changes}{
violate change break enforce follow use challenge practice write obey adopt impose revise draft apply teach reform amend ignore uphold
}
\R{v}{163}{0.11}{political consideration}{
pass support approve oppose vote sign introduce propose veto consider reach include block sponsor favor want reject back push get % (0.678 mass)
}
\R{v}{10}{0.04}{judicial review}{
  apply permit challenge give interpret bar strike impose limit involve recognize uphold OOV adopt justify regulate seek place define hold
}
\R{v}{241}{0.04}{generic/light verbs}{
  have OOV do find take expect see allow use create begin view produce place suffer add like start study face
}
\R{s}{71}{0.60}{jurisdictions}{
  state Congress government system OOV court York city judge law county Jersey Government Legislature State official California legislator country States
}
\R{s}{188}{0.14}{legislative and executive branches}{
  Senate Legislature House Congress Republicans lawmaker Pataki governor Clinton Bush assembly Democrats OOV leader legislator Administration Cuomo Council administration group
}
\R{o}{42}{0.80}{kinds/aspects of laws, regulations}{
  law statute rule guideline order system regulation ordinance Constitution curfew Act ban restriction policy code provision legislation requirement agreement limit
}
\R{o}{332}{0.09}{(same as above)}{
  penalty discretion parole use number limit period punishment system power OOV approval sale provision test type time judge release protection
}
\R{o}{395}{0.06}{procedural terms}{
  bill measure legislation amendment provision proposal law ban version vote package penalty veto Senate agreement issue abortion language passage action
}
\frameend

\caption{Example frame (``Legislation'') learned from \textsc{CrimeNYT}.}
\label{legislation}
\end{table}

Every word class, of course, can be used multiple times by several different frames.
We were curious if the model could find word classes that took the subject position
in some frames, but the object position in others---\cite{chambers2009unsupervised}
demonstrate interesting examples of this in their learned schemas.
Our model does find such instances.
Consider class $c=99$, a ``victim'' class: \emph{\{girl boy student teen-ager daughter victim sister child\}}.
It appears as an object for a ``violence/abuse against victim'' frame
where the most common verb class has top words
\emph{\{rape assault attack meet force identify see ask\}},
while it is a subject in a frame with a wider variety
of generic and communication verbs \emph{\{say be go come try admit ask work agree continue refuse\}}.

% \begin{table}
%   \footnotesize
% \fframestart{159}{}
% \ffr{v}{98}{0.21}{
% say be go come try admit ask work agree continue refuse write plead recall live seem insist sit suggest acknowledge
% }
% \ffr{v}{199}{0.18}{
%   testify say tell describe complain appear ask stand die reply move be cry lie sit act decide weep disappear recount
% }
% \ffr{s}{\textbf{99}}{0.71}{
%   girl boy student teen-ager daughter victim sister child worker neighbor teenager patient classmate woman mother teacher adult breast youth time
% }
% % \ffr{s}{339}{0.20}{
% %   mother father family son friend parent brother daughter wife husband sister relative neighbor member couple Katie girlfriend grandmother Esposito aunt
% % }
% % \ffr{o}{131}{0.10}{
% % anything man time question statement charge friend day word detail office call note request lawyer year letter testimony head home}
% \fframeend
% 
% \fframestart{119}{}
% \ffr{v}{250}{0.47}{
%   rape assault attack meet force identify see ask sodomize abuse lure touch pick fondle treat grope kidnap OOV grab involve
% }
% \ffr{v}{292}{0.17}{
%   abuse molest protect involve take touch assault fondle have place rape adopt sodomize interview exploit lure grope educate send remove
% }
% \ffr{o}{\textbf{99}}{0.91}{
%   girl boy student teen-ager daughter victim sister child worker neighbor teenager patient classmate woman mother teacher adult breast youth time
% }
% \fframeend
% \caption{Frames (partial) that use ``victim/family'' class $c=99$.}
% \label{victim}
% \end{table}

\bocomment{word senses: attack}

\bocomment{word senses\ldots sodomize? yikes.}

There are many more interesting examples.  A sampling of other word classes, with our interpretations, include:

\begin{itemizesquish}
  \item c=3 \syncolor{v}{(v)} stages in a process, esp. criminal process:
  \emph{serve face receive await stand enter spend complete accept get violate avoid post deny reduce give finish begin grant draw revoke jump}
\item
  c=4\syncolor{v}{(v)} argumentation: \emph{prove demonstrate avoid reflect reach undermine carry mean affect force satisfy}; with the message/argument as subject: (c=67) \emph{\{case decision arrest ruling action\}} and (c=120) \emph{\{issue view race theme record kind promise\}}
\item
  c=16\syncolor{v}{(v)} physical pursuit and apprehension/confrontation:
\emph{force approach try flee follow threaten know drag tell escape take find walk hold knock grab order push admit break describe leave climb}
\item
  c=17\syncolor{o}{(obj)} effects:
\emph{effect impact role consequence anything implication case value connection importance link interest root basis significance bearing}
\item
  c=19\syncolor{o}{(obj)}: verdicts/sentences:
\emph{sentence penalty punishment execution term order conviction leniency life fine clemency verdict sentencing factor date death circumstance}
\item
  c=44\syncolor{o}{(obj)} games \& pastimes:
\emph{role game part OOV basketball card ball football music song baseball sport host piano soccer golf tape politics guitar tennis bottom season}
\item
  c=46 (\textcolor{ForestGreen}{subj},\textcolor{blue}{obj}) societal ills:
\emph{problem violence crime issue situation activity abuse behavior flow criminal kind fear tide cause corruption spread threat case root type crisis}
\end{itemizesquish}

\noindent
To guard against a cherry-picking bias,
we include an easy-to-read report of all
frames, classes, and linkings
online.\footnote{\url{http://brenocon.com/dap/materials/}}

Some classes are very general, and some are very specific.
One interesting thing to note is that some classes have a more topical, and less syntactically coherent,
flavor.  
% Some frames can lump together words that do not necessarily work in their argument positions,
% but are important for the broader topic or theme; 
For example, c=18: \emph{\{film viewer program network movie show producer station audience CBS television camera actor fan\}}.  It appears often for only one frame,
and is split 2:1 between subject and object position.  
Essentially, the syntactic positioning is being ignored:
$p(c|f,a=subj)$ is only twice as likely as $p(c|f,a=obj)$, 
whereas for most noun classes this ratio is in the hundreds or thousands.
This word class functions more like an LDA topic.
Is it appropriate to interpret it as a ``topic,''
or does it correspond to entities active in
a Fillmorean-style frame of ``Television Show''?
By leveraging both document and syntactic context,
we believe our model uncovers semantics for situations and events.

\bocomment{near-duplicates, like prosecutors and say \dots i blame the symmetric prior}

\section{Comparison to FrameNet} \label{se:fn}

We are primarily interested in the quality of our induced frame lexicon.
Evaluation is difficult;
one automatic measure, held-out likelihood,
may not always correlate to subjective semantic coherency \citep{chang_reading_2009}.
And while subjective coherency judgments are
often collected
to evaluate word clusters or argument compatibilities,
it is unclear to us exactly what task setup would directly support analyzing frame quality.
Our primary goal is to achieve a better understanding of what our model is and is not learning.

We propose a method to compare the similarities and differences
between a learned frame model and a pre-existing lexicon.
\cite{chambers_template-based_2011}
compare their learned frames to MUC templates in the domain of news reports
about terrorist activities.
Seeking a resource that is more general, more lexicon-focused, 
can be used to compare different
corpus domains
we turn to FrameNet \citep{fillmore2001frame,ruppenhofer2006framenet},
% (CITE),\footnote{URL}
a well-documented
lexical resource of actions/situations and their typical participant types.
In this section we present
 an analysis of
 wordset-to-wordset similarity alignments, that we use to analyze verb clusters.
 % ; and
% two analyses, based on
% \begin{enumerate}
%   \item
%  Wordset-to-wordset similarity alignments, that we use to analyze verb clusters; and
%  \item
%  Frame-to-frame similarity alignments, that analyze more complex frame-role-wordset structures.
%  \bocomment{not done yet\ldots will pluck out a few examples I guess.  comprehensive analysis across everything is too much if we do that for the verbs} \bocomment{gonna cut soon\ldots}
% \end{enumerate}

% /d/lexical/framenet/fndata-1.5 % grep '<frame' frameIndex.xml | wc -l
%     1020

The downloadable FrameNet 1.5 dataset consists of 1,020 \emph{frames}, each of which is associated with a number of \emph{lexical units}, that, when used,
% as a \emph{target} word,
% \footnote{
%   Some researchers call this a \emph{predicate} (Das and Smith, Das et al. \ldots do other FN SRL papers use this terminology?)} 
can evoke their respective frames.  
A lexical unit is essentially a word sense; it is associated with a single frame, and also a \emph{lexeme}, which consists of a word \emph{lemma} and part-of-speech category.
In the FrameNet annotations, the frame-evoking word is known as the \emph{target}; some of the frame's roles (called \emph{frame elements}) then are bound to words and phrases in the annotated sentence.
\bocomment{gonna need an example, that's a lot of terminology. and/or cut it down}

% /d/lexical/framenet/fndata-1.5 % grep '<lu' luIndex.xml | grep -Po 'name="[^"]*"'|grep -Po '\.[^\.]*"$'|count|sort -n
% 3 .intj"
% 3 .scon"
% 4 .art"
% 6 .c"
% 33 .num"
% 143 .prep"
% 167 .adv"
% 2122 .a"
% 4605 .v"
% 4743 .n"

Most of FN's lexical units are verbs, nouns, or adjectives.
In this work we
focus on the verbs, which in FN's annotated data
often take their grammatical subject and object
as frame arguments.  
FN may not have been constructed with this purpose in mind,
but by viewing FN as a database of verb clusters and typical argument types,
it has a structure comparable to our Model 1, and, to a lesser extent, Model 2.
And while FrameNet was constructed to go beyond just verbs---in contrast to another similar frame resource,
VerbNet---we find that extracting a verb and argument clusters from FN in this way
yields a reasonable-looking dataset.

% /d/lexical/framenet/fndata-1.5 % grep '<lu' luIndex.xml |grep 'hasAnnotation="true"' | wc -l
%     7711
% /d/lexical/framenet/fndata-1.5 % grep '<lu' luIndex.xml |grep -v 'hasAnnotation="true"' | wc -l
%     4119
% /d/lexical/framenet/fndata-1.5 % grep '<lu' luIndex.xml |wc -l
%    11830
% /d/lexical/framenet/fndata-1.5 % grep '<lu' luIndex.xml | grep -Po 'name="[^"]*"'|grep '\.v"'|wc -l
%     4605

FN has 11,830 lexical units, 4,605 of which are verbs.
We group lexical units by frame, and filter to frames that have 5 or more verbs
as lexical units.  This leaves 171 frames with 1829 unique verb lemmas.  
(We ignore multiword lexemes as well as non-verb lexemes.)

\subsection{Comparing verb wordsets}
How can we analyze the similarity and differences of two different word clusterings?
Our notion of ``clustering'' need not be a proper partition:
it merely consists of some number of \emph{wordsets},
which may overlap and not necessarily cover the entire vocabulary.
(These are often referred to as ``word clusters,'' but for clarity we always use ``wordset.'')
Many lexical resources and unsupervised models can be discretized and converted
into this representation.
% (CITE the verb cluster eval stuff)

\begin{figure}\centering
  \includegraphics[width=1.8in]{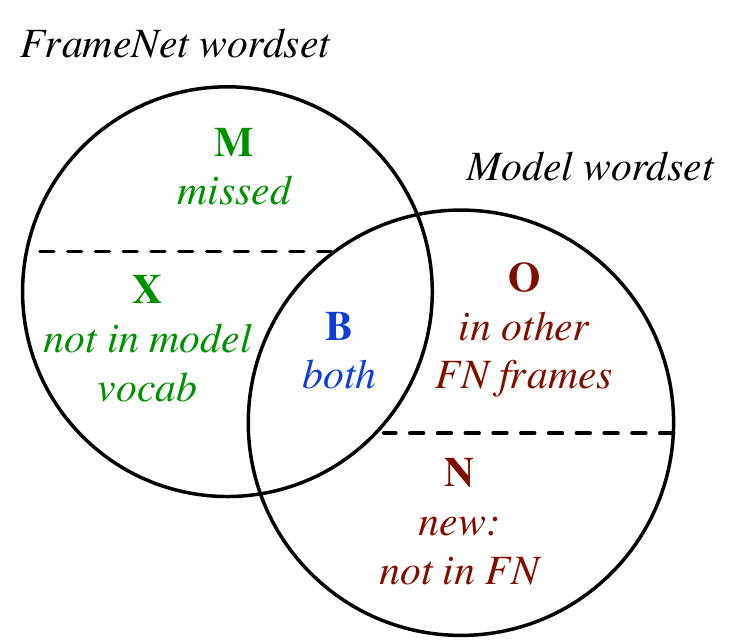}
\caption{Quantities in a Venn comparison of two wordsets, as used in Table~\ref{tab:fnmatch}.
Note ``N'' actually means it appears fewer than 5 times in the
FrameNet annotated data.
% \bocomment{TODO $<5$ fix re nathan}
}
\label{venndiagram}
\end{figure}

We perform a basic analysis, comparing the verb wordsets implied by FrameNet to our model.
Verb wordsets (verbsets) are extracted from FrameNet by taking the set of verbs for each frame, so there are 171 verbsets.  
We discretize our model by taking,
for every word class,
the words having a count of at least 5 in the Gibbs sample being analyzed.
% (In principle, we might want to use all words with non-zero counts, or having a substially high posterior
% probability for the class.  
We observed that words with smaller counts than this
tended to be unrelated or marginally related to the others---their presence
may be due to the randomness inherent in any single Gibbs sample.

The similarity measures are as follows.  For two wordsets $A$ and $B$, the Dice coefficient 
%(CITE)
is
\[ DiceSim(A,B) = \frac{2 |A\cap B|}{|A| + |B|} \]

\noindent
This is in fact equivalent to the F1-measure, and monotonic in Jaccard similarity; 
see the appendix for details.
% see our technical report (CITE .. unless Appendix) for details.
Let $A_i$ be the FrameNet verbset for FN frame $i$,
and $B_j$ be a model's verbset.
We compute all $DiceSim(A_i,B_j)$ similarities, and for each FN frame, find the best model match $j$,
\[ \arg\max_j DiceSim(A_i, B_j) \]
and show the best matches in Table \ref{tab:fnmatch}.
This is done with \textsc{CrimeNYT} with 400 frames and classes.

The best match, for FrameNet's ``Change position on a scale,'' 
clearly has a common semantic theme shared between the FrameNet and model verbsets.
The model fails to get several words such as ``decline,''
but does have several other words that seem plausible candidates to add to FrameNet
here: ``plunge,'' ``sink,'' ``surge.''

We show in Table \ref{multitable} the matches against the different corpora.
Interestingly, different corpora are better at recovering different types of frames.

One big issue with this metric is that FrameNet is not designed to have complete lexical coverage for a particular role, so it is unfair to penalize our model for learning novel words not in FrameNet for a particular role.  On the other hand, some of the learned new words are sometimes clearly not semantically related.  When a model's wordset scores are low, we don't know whether it's because it's actually semantically incoherent, or if FrameNet had poor coverage in its area.  It is important to analyze specific instances that make up the quality measure, as in Table \ref{tab:fnmatch}.

Our method may be more useful as part of a semi-automated system to suggest new additions to a resource like FrameNet; to do this well, it may be interesting to explore building supervision in to an earlier stage of the model, rather than in posthoc analysis as we develop it here.  One possibility is to use the data as infomed priors: have asymetric Dirichlet priors with higher values for roles seen in the FrameNet data.

\newcommand{\leftnamebox}[1]{\parbox{0.50in}{#1}}
\newcommand{\bothbox}[2] {\parbox[c]{1.15in}{#1 #2}}
\newcommand{\blabox}[4]{ 
  \parbox{1.9in}{
  #1 #2 \\ #3 #4
  }
}
\newcommand{\fonlybox}[4]{\blabox{#1}{#2}{#3}{#4}}
\newcommand{\monlybox}[4]{\blabox{#1}{#2}{#3}{#4}}
\newcommand{\fnmatchthresh}[0]{0.15}
\newcommand{\cc}[1]{ {\tiny \emph{[#1]}} }
\newcommand{\countboth}[1]{    \cc{B={#1}} }
\newcommand{\countfmissed}[1]{ \cc{M={#1}} }
\newcommand{\countfoov}[1]{    \cc{X={#1}} }
\newcommand{\countmmissed}[1]{ \cc{O={#1}} }
\newcommand{\countmoov}[1]{    \cc{N={#1}} }
\begin{table*}\footnotesize\centering
  % \begin{tabular}{|p{0.55in}|p{0.27in}|p{1.4in}|p{1.4in}|p{1.4in}|}
  \begin{tabular}{|l|l|l|l|l|}
    \hline
    FN frame & Dice & In both & Only in FN & Only in model\\
    \hline
\leftnamebox{ Change position on a scale }  &              0.320  &              \bothbox{ plummet, skyrocket, tumble, dwindle, double, rise, triple, fall }{ \countboth{12} }  &              \fonlybox                 { decline, rocket, mushroom, advance, drop, reach }{ \countfmissed{16} }                 {  }{\countfoov{0} }  &              \monlybox                 { shoot, represent, plunge, return, appear, show }{ \countmmissed{16} }                 { be, exceed, hover, sink, stabilize, surge }{    \countmoov{19} }          \\ \hline
\leftnamebox{ Statement }  &              0.265  &              \bothbox{ assert, suggest, explain, add, note, say, caution, report }{ \countboth{9} }  &              \fonlybox                 { comment, attest, relate, address, insist, allege }{ \countfmissed{29} }                 { aver, pout, conjecture, avow, gloat }{\countfoov{5} }  &              \monlybox                 { respond, decline, appear, describe, testify, indicate }{ \countmmissed{10} }                 { be, deny, refuse, continue, emphasize, refer }{    \countmoov{6} }          \\ \hline
\leftnamebox{ Text creation }  &              0.227  &              \bothbox{ write, type, pen, compose, draft }{ \countboth{5} }  &              \fonlybox                 { utter, say, chronicle, author }{ \countfmissed{4} }                 { jot }{\countfoov{1} }  &              \monlybox                 { translate, prepare, get, read, study, contribute }{ \countmmissed{14} }                 { promote, edit, censor, deliver, submit, research }{    \countmoov{15} }          \\ \hline
\leftnamebox{ Request }  &              0.218  &              \bothbox{ urge, beg, summon, implore, ask, tell }{ \countboth{6} }  &              \fonlybox                 { command, demand, request, order, plead }{ \countfmissed{5} }                 { entreat, beseech }{\countfoov{2} }  &              \monlybox                 { give, visit, telephone, notify, thank, lead }{ \countmmissed{22} }                 { warn, convince, invite, quote, defy, cross-examine }{    \countmoov{14} }          \\ \hline
\leftnamebox{ Killing }  &              0.206  &              \bothbox{ starve, murder, slaughter, drown, kill, lynch, assassinate }{ \countboth{7} }  &              \fonlybox                 { slay, liquidate, massacre, smother, butcher, dispatch }{ \countfmissed{8} }                 { crucify, asphyxiate, annihilate, garrotte, behead, decapitate }{\countfoov{6} }  &              \monlybox                 { blow, shoot, hit, torture, injure, intimidate }{ \countmmissed{18} }                 { harm, cheat, guard, strangle, kidnap, dismember }{    \countmoov{22} }          \\ \hline
\leftnamebox{ Evidence }  &              0.200  &              \bothbox{ reveal, show, contradict, support, prove, confirm, indicate, demonstrate }{ \countboth{8} }  &              \fonlybox                 { attest, verify, testify, evidence, corroborate, disprove }{ \countfmissed{9} }                 { evince }{\countfoov{1} }  &              \monlybox                 { emerge, conclude, relate, describe, discover, examine }{ \countmmissed{25} }                 { point, focus, imply, exist, result, determine }{    \countmoov{29} }          \\ \hline
\leftnamebox{ Compliance }  &              0.185  &              \bothbox{ violate, flout, break, observe, follow, obey }{ \countboth{6} }  &              \fonlybox                 { conform, breach, comply, adhere }{ \countfmissed{4} }                 { contravene }{\countfoov{1} }  &              \monlybox                 { use, set, contradict, ease, evade, soften }{ \countmmissed{14} }                 { loosen, adopt, rewrite, strengthen, revamp, administer }{    \countmoov{34} }          \\ \hline
\leftnamebox{ Getting }  &              0.182  &              \bothbox{ win, acquire, obtain, gain }{ \countboth{4} }  &              \fonlybox                 { get }{ \countfmissed{1} }                 {  }{\countfoov{0} }  &              \monlybox                 { secure, await, terminate, want, demand, seek }{ \countmmissed{6} }                 { owe, dole, trade, need, withhold, guarantee }{    \countmoov{29} }          \\ \hline
\leftnamebox{ Experiencer obj }  &              0.176  &              \bothbox{ satisfy, shock, offend, infuriate, puzzle, reassure, scare, enrage }{ \countboth{17} }  &              \fonlybox                 { unsettle, distress, rattle, frighten, confuse, sting }{ \countfmissed{48} }                 { mortify, displease, exhilarate, disconcert, astound, hearten }{\countfoov{48} }  &              \monlybox                 { despise, love, divide, back, bring, want }{ \countmmissed{24} }                 { force, transfix, owe, haunt, involve, persuade }{    \countmoov{39} }          \\ \hline
\leftnamebox{ Hindering }  &              0.170  &              \bothbox{ hinder, obstruct, impede, hamper }{ \countboth{4} }  &              \fonlybox                 { inhibit, interfere }{ \countfmissed{2} }                 { trammel, constrain, encumber }{\countfoov{3} }  &              \monlybox                 { head, thwart, lead, avoid, pass, harass }{ \countmmissed{16} }                 { insure, overstep, harm, suspend, monitor, intercept }{    \countmoov{18} }          \\ \hline
\leftnamebox{ Awareness }  &              0.167  &              \bothbox{ conceive, suspect, know }{ \countboth{3} }  &              \fonlybox                 { comprehend, understand, presume, imagine, believe, think }{ \countfmissed{6} }                 { reckon }{\countfoov{1} }  &              \monlybox                 { notice, calm, stereotype, rattle, recognize, like }{ \countmmissed{14} }                 { dig, remember, figure, spell, prefer, suppose }{    \countmoov{9} }          \\ \hline
\leftnamebox{ Telling }  &              0.163  &              \bothbox{ inform, tell, notify, assure }{ \countboth{4} }  &              \fonlybox                 { confide, advise }{ \countfmissed{2} }                 { apprise }{\countfoov{1} }  &              \monlybox                 { give, question, telephone, thank, lead, visit }{ \countmmissed{24} }                 { warn, convince, invite, remind, quote, defy }{    \countmoov{14} }          \\ \hline
\leftnamebox{ Activity start }  &              0.162  &              \bothbox{ start, begin, enter }{ \countboth{3} }  &              \fonlybox                 { initiate, launch, commence, swing }{ \countfmissed{4} }                 {  }{\countfoov{0} }  &              \monlybox                 { quit, cut, run, attend, skip, fix }{ \countmmissed{11} }                 { finish, complete, reform, disrupt, rock, offer }{    \countmoov{16} }          \\ \hline
\leftnamebox{ Path shape }  &              0.160  &              \bothbox{ dip, drop, reach, edge }{ \countboth{4} }  &              \fonlybox                 { emerge, swerve, angle, veer, crisscross, snake }{ \countfmissed{17} }                 { traverse, ascend, undulate, slant, zigzag, ford }{\countfoov{7} }  &              \monlybox                 { jump, accord, move, begin, stay, explode }{ \countmmissed{10} }                 { slow, figure, rebound, tend, constitute, range }{    \countmoov{8} }          \\ \hline
\leftnamebox{ Cause harm }  &              0.157  &              \bothbox{ hit, torture, injure, bludgeon, hurt, stab, strike, batter }{ \countboth{8} }  &              \fonlybox                 { cane, pummel, bruise, clout, hammer, whip }{ \countfmissed{36} }                 { electrocute, spear, pelt, cudgel, squash, horsewhip }{\countfoov{11} }  &              \monlybox                 { blow, starve, murder, intimidate, impress, ambush }{ \countmmissed{17} }                 { harm, cheat, guard, strangle, kidnap, dismember }{    \countmoov{22} }          \\ \hline

\hline
  \end{tabular}
  \caption{FN verbset single best matches to Model 2 on \textsc{CrimeNYT}, $F=C=400$, 
  having a best-Dice score at least \fnmatchthresh.
  We break down the set comparison as per Figure~\ref{venndiagram},
  showing up to several words words from each subset.
  The full set size is given in square brackets.
  {\footnotesize
  $B=$ in both wordsets.
  $M=$ ``missed'': not in this model wordset.
  $X=$ not in model's vocabulary.
  $O=$ in other frames.
  $N=$ ``new'': not in our FN verb extraction, i.e. appears fewer than 5 times 
  in the FrameNet annotated data.
  % Dice score is $2B / (2B+M+X+O+N)$, same as F-measure.
  % \bocomment{do nathan's $<5$ fix}
  }
}
  \label{tab:fnmatch}
\end{table*}

\begin{table}\centering\scriptsize
% \begin{minipage}[t]{0.5\linewidth}
  \vspace{-3.1em}
  \begin{tabular}{|l|ccccccc|}
  \hline FN frame & \rotatebox{90}{BrownPTB (100)} & \rotatebox{90}{BrownPTB (400)} & \rotatebox{90}{WSJPTB (100)} & \rotatebox{90}{WSJPTB (400)} & \rotatebox{90}{CrimeNYT (100)} & \rotatebox{90}{CrimeNYT (400)} & \rotatebox{90}{UnifNYT (400)} \\
\hline
Change pos.~on a scale & 10 & 10 & \textbf{57} & \textbf{49} & 18 & \textbf{32} & \textbf{37} \\
Statement & 11 &  & \textbf{31} & \textbf{26} & \textbf{34} & \textbf{26} & \textbf{28} \\
Cause chg. of pos.~on scl. & 14 & 13 & 18 & \textbf{26} &  & 11 & 12 \\
Body movement & 18 & 15 & 10 &  & 12 & 14 & \textbf{21} \\
Awareness &  & \textbf{27} &  & 11 &  & 16 & 15 \\
Motion directional &  &  & \textbf{23} & 18 &  & 14 & 13 \\
Appearance & 14 & 17 & 10 &  &  & 14 & 14 \\
Becoming & 15 & 12 &  & 10 &  & 13 & 16 \\
Evidence &  &  &  & 12 &  & 20 & 19 \\
Arriving &  & 13 &  & 11 &  & 12 & \textbf{21} \\
Causation &  & 14 &  & \textbf{22} &  & 12 & 10 \\
Using & 18 & 20 &  & 16 &  &  &  \\
Path shape &  &  & 15 & 16 &  & 16 & 12 \\
Getting &  &  &  & 13 &  & 18 & 15 \\
Cause harm & 14 &  &  &  & 11 & 15 & 16 \\
Self motion & 10 &  &  &  & 13 & 12 & 15 \\
Motion &  &  & 13 & 16 &  &  & 14 \\
Request &  &  &  &  &  & \textbf{21} & 18 \\
Change posture &  & 12 & 10 &  &  &  & 12 \\
Becoming aware & 10 & 11 &  & 12 &  & 11 & 12 \\
Coming to be & 18 &  &  & 12 &  & 12 &  \\
Cause change &  & 10 &  & 14 &  & 11 & 15 \\
Manipulation &  & 10 &  &  & 13 & 12 & 13 \\
Ingest substance & 11 & 12 &  &  &  &  & 11 \\
Compliance &  &  &  &  &  & 18 & 17 \\
Coming to believe &  &  &  & 16 &  & 11 & 10 \\
Perception experience & 13 &  &  & 12 &  & 11 &  \\
Departing & 14 &  &  &  &  & 11 &  \\
Removing &  &  &  &  &  & 13 & 12 \\
Contingency &  & 16 &  &  &  &  & 15 \\
Cotheme &  &  &  & 16 &  & 10 & 13 \\
Bringing &  &  &  & 12 &  & 13 & 10 \\
Categorization &  & 14 &  & 11 &  &  &  \\
Topic &  &  &  & 10 &  & 14 & 15 \\
Placing &  &  &  &  & 10 & 12 & 10 \\
Communicate categ'n &  &  &  & 13 &  & 12 & 10 \\
Reveal secret & 15 &  &  &  &  & 12 & 12 \\
Activity start &  &  &  & 11 &  & 16 &  \\
Text creation &  &  &  &  &  & \textbf{22} &  \\
Expectation &  & 11 &  & 10 &  &  &  \\
Experience bodily harm & 11 &  &  &  &  & 10 &  \\
Cause motion & 11 &  &  &  &  &  &  \\
Leadership &  & 14 &  & 10 &  &  & 10 \\
Experiencer obj &  &  &  &  & 11 & 17 & 12 \\
Cause expansion & 13 &  &  & 15 &  &  &  \\
Posture & 10 &  &  &  &  & 10 & 11 \\
Perception active &  &  &  &  &  & 10 & 10 \\
Birth &  &  & 11 & 10 &  &  & 10 \\
Cooking creation &  &  &  & 14 &  & 10 &  \\
Activity stop & 16 &  &  &  &  &  & 11 \\
Traversing &  &  &  &  &  &  & 11 \\
Giving &  &  &  &  &  & 12 & 11 \\
Ride vehicle &  &  &  &  &  &  & 18 \\
Building &  &  & 13 & 12 &  &  &  \\
Contacting &  &  &  &  &  & 14 & 11 \\
Filling &  &  &  &  &  & 10 &  \\
Killing &  &  &  &  &  & \textbf{20} & 15 \\
Telling &  &  &  &  &  & 16 & 16 \\
Cause impact &  &  &  &  &  &  &  \\
Reasoning &  &  &  & 12 &  & 11 &  \\
 \hline
  \end{tabular}
% \end{minipage}
% \begin{minipage}[t]{0.5\linewidth}
%   \begin{tabular}{|l|ccccccc|}
%   \hline \input{emore/multitable2.tex} \hline
%   \end{tabular}
% \end{minipage}
  \caption{\footnotesize For a FrameNet frame's verbset, its single-best-match Dice scores (multiplied by 100)
    against several different models and datasets.
  Scores less than 0.1 are not shown; greater than 0.2 are bolded.
  60 frames are shown, in decreasing order of average match score across datasets/models.  
  All runs are Model 2, with $F=C=$ the number in parentheses.
  % \bocomment{maybe go to just one column, 50 or 60 rows or so}
}
  \label{multitable}
\end{table}

\bocomment{do FN matching scores stabilize early in the training process?  or get worse over time? how closely correlate to likelihood?}

\bocomment{look at model frames with no good match to FN frames}

\bocomment{noun classes are some nice easy ones here, like illegal drugs}

% \subsection{Aggregate matching scores}
% 
% Furthermore, we can aggregate these into a full score
% for how well 
% FrameNet is matched by the learned model---i.e., how well it has been reconstructed---as the average of the single-best-match alignments,
% 
% \[ FNMatchScore = \frac{1}{N} \sum_i \max_j DiceSim(A_i, B_j) \]
% 
% % And conversely, how well FrameNet provides a match for the model,
% % \[ ModelMatchScore = \frac{1}{M} \sum_j \max_i DiceSim(A_i, B_j) \]
% 
% This bears some resemblance to Lang and Lapata's use of purity and collocation measures for evaluation (CITEYEAR),
% \bocomment{and cite new/older ones we just found}
% which average the best intersection sizes instead of Dice scores.
% We prefer Dice (F-measure), since it seems to correspond better to the
% qualitative semantic similarity between two wordsets, and we would like our
% aggregated score to be a direct combination of the similarity scores we use for
% drilldown analysis.
% 
% \bocomment{table of them for all conditions }
% 
% 
% \subsection{Comparing frame structures}
% 
% We use a similar pair-scoring approach to analyze fuller frame structures of FrameNet against our model.
% This is substantially tricker than comparing wordsets,
% since a frame has substantial substructure, and our model's notion of frames
% is a bit different than FrameNet's.
% 
% \bocomment{tricky\ldots average max scores \ldots}
% 
% \bocomment{also it's SLOW}

% \section{Related Work}
% bla

\section{Note on MUC}

Besides FrameNet, it may be worth comparing to a verb clustering
more like the \cite{levin1993english} classes;
for example, \cite{sun2008verb} and \cite{sun2009improving}
construct a set of 204 verbs in Levin-style clusters
and evaluate clustering methods against them.

It would be interesting to compare to the work of
\cite{chambers11}, which induces frames with 
several stages of ad-hoc clustering on unlabeled newswire data,
and compares its learned frames
to frames and extractions from MUC-4, a domain of newswire reports of terrorist
and political events.

We did perform a headroom test for MUC extraction accuracy and found our approach is 
poor for this task---by looking at all role-filler 
instances in the text,
and checking how often they corresponded to a nominal subject or object according to the Stanford dependency parser plus our syntactic extraction rules.  This was 42\% of the time (on the \textsc{DEV} data), which constitutes an upper bound on recall.  
Many of the MUC instances (as well as FrameNet annotations)
use noun-noun, implicit, and other syntactic indicators of semantic relations.
The approach of \cite{Cheung2013Frames} is superior for this task:
they use a flexible set of dependency relations integrated with discourse-level
sequence model, and compare favorably to \citeauthor{chambers11}'s work.

(In the process of our analysis we did a clean-up of the original data
that makes it easier to use, and we have made freely available for futher
research.\footnote{\url{http://brenocon.com/muc4_proc/} and \url{http://github.com/brendano/muc4_proc}}
Unfortunately, due to the complexity of MUC data, there is
some
ambiguity on how to conduct an evaluation.  We
uncovered a number of discrepancies between 
the evaluation done by
\citeauthor{chambers11}
versus the previous work they compare to
\citep{pat07}; after a number of email exchanges with all previous co-authors,
Chambers modified his evaluation implementation and reports minor changes to 
their evaluation numbers.
We have assembled a document with evaluation methodology clarifications
from Chambers, Patwardhan, and Riloff, and posted it online.\footnote{\url{https://docs.google.com/document/pub?id=1erEEsWI9V0SapEecbn1AMy69Fy6TgSIdYVTsKRaF8vM}
})

\section{Conclusion}

We have illustrated a probabilistic model that learns frames from text,
combining document and syntactic contexts in a Dirichlet-multinomial latent variable model.  Many further extensions are possible.  First, document context could be enriched with various metadata---a document's context in time, space, and author attributes can easily be incorporated in the graphical models framework.  Second, the restrictions to verb-subject-object syntactic constructions must be relaxed in order to capture the types of arguments seen in semantic role labeling and information extraction.

\section*{Acknowledgments}

Thanks to Noah Smith and Nathan Schneider for advice and assistance in model analysis.

\section{Appendix}

\subsection{Dice, F-measure, and set similarity}\footnote{A version of this appendix was published at
\url{http://brenocon.com/blog/2012/04/f-scores-dice-and-jaccard-set-similarity/}.}
Let $A$ be the set of found items, and $B$ the set of wanted items.
$Prec=|AB|/|A|$, $Rec=|AB|/|B|$.  Their harmonic mean, the $F1$-measure,
is the same as the Dice coefficient:
\begin{align*}
F1(A,B) 
&= \frac{2}{1/P+ 1/R} 
 = \frac{2}{|A|/|AB| + |B|/|AB|} \\
Dice(A,B) 
&= \frac{2|AB|}{ |A| + |B| } \\
&= \frac{2 |AB|}{ (|AB| + |A \setminus B|) + (|AB| + |B \setminus A|)} \\
% Tversky(A,B;\alpha=\frac{1}{2},\beta=\frac{1}{2})
&= \frac{|AB|}{|AB| + \frac{1}{2}|A \setminus B| + \frac{1}{2} |B \setminus A|}
\end{align*}
This illustrates Dice's close relationship to the Jaccard metric,
\begin{align*}
Jacc(A,B) 
&= \frac{|AB|}{|A \cup B|} \\
&= \frac{|AB|}{|AB| + |A \setminus B| + |B \setminus A|}
\end{align*}
And in fact $J = D/(2-D)$ and $D=2J/(1+J)$ for any input, so they are monotonic in one another.  
The Tversky index (1977) generalizes them both,
\begin{align*}
Tversky(A,B;\alpha,\beta) 
&= \frac{|AB|}{|AB| + \alpha|A\setminus B| + \beta|B \setminus A|}
\end{align*}
where $\alpha$ and $\beta$ control the magnitude of penalties of false positive versus false negative errors.
All weighted $F$-measures correspond to when $\alpha+\beta=1$.

\subsection{Appendix: Dirichlet-multinomial conjugacy and the DM}
\label{dcm_math}

Consider the two-stage model
\[ \theta\sim Dir(\alpha),\ \ x\sim Multinom(\theta) \]

\noindent
where $\alpha$ is a real-valued vector Dirichlet parameter, and $x$ is a vector of outcome counts.
Let $A=\sum_k \alpha_k$; this is the concentration parameter.

$p(x|\alpha) = \int p(x|\theta) p(\theta|\alpha) d\theta$ is the Dirichlet-multinomial, a.k.a.~Multivariate Polya distribution or Dirichlet-compound multinomial.   It is a distribution over count vectors,
just like the multinomial, except it has the capacity to prefer different levels of sparseness vs. non-sparseness.

\noindent First note the Dirichlet density
\begin{align}
p(\theta|\alpha) 
&= \frac{\Gamma A}{\prod\Gamma\alpha_k} \prod \theta_k^{\alpha_k-1} 
\ \ =\ \ \frac{1}{B(\alpha)} \prod \theta_k^{\alpha_k-1}
\end{align}
where $B(\alpha)$ is the multivariate Beta function
\[ B(\alpha) = \int \prod \theta_k^{\alpha_k-1} d\theta = \frac{\prod\Gamma \alpha_k}{\Gamma A} \]

\noindent Now derive the DM PMF:
\begin{align}
p(x|\alpha) 
&= \int p(x|\theta)\ p(\theta|\alpha)\ d\theta \\
&= \int Multinom(x;\theta)\ Dir(\theta;\alpha)\ d\theta \\
&= \int 
  \left(\frac{N!}{\prod x_k!} \prod\theta_k^{x_k} \right)
  \left(\frac{1}{B(\alpha)}\prod\theta_k^{\alpha_k-1} \right)
d\theta
\end{align}

\noindent
Because of conjugacy (i.e., the densities play nicely under multiplication),
we can combine them into the
integral of a new unnormalized Dirichlet, and rewrite into closed form.

\begin{align}
p(x|\alpha)
&= \frac{N!}{\prod x_k!}\frac{1}{B(\alpha)}
  \int 
  \prod
    \theta_k^{x_k + \alpha_k - 1} d\theta \\
&= \frac{N!}{\prod x_k!}
\frac{B(\alpha+x)}{B(\alpha)}
\\
DM(x; \alpha) &\equiv 
\underbrace{
  \frac{N!}{\prod x_k!} }_{
  \text{n.~seq} 
}
\underbrace{
  \frac{\Gamma(A)}{\Gamma(A+N)}
  \prod\frac{\Gamma(\alpha_k+x_k)}{\Gamma(\alpha_k)}
}_{ 
  \text{prob of a seq having counts $\vec{x}$}
}
\end{align}
where $A=\sum\alpha_k$ and $N=\sum x_k$. 
To calculate this, one would rewrite the ``number of sequences term'' using
$N!=\Gamma(N+1)$ and $x_k!=\Gamma(x_k+1)$ then use the log-gamma function for everything.
To calculate the log-probability of only a single sequence, omit the initial term
$N!/\prod x_k!$; we call this a ``single path DM'' or ``DM1'':

\begin{align} DM1(x; \alpha) &= \
\frac{\Gamma(A)}{\Gamma(A+N)}
\prod\frac{\Gamma(\alpha_k+x_k)}{\Gamma(\alpha_k)}
\end{align}

\begin{figure}\centering
\includegraphics[width=4in]{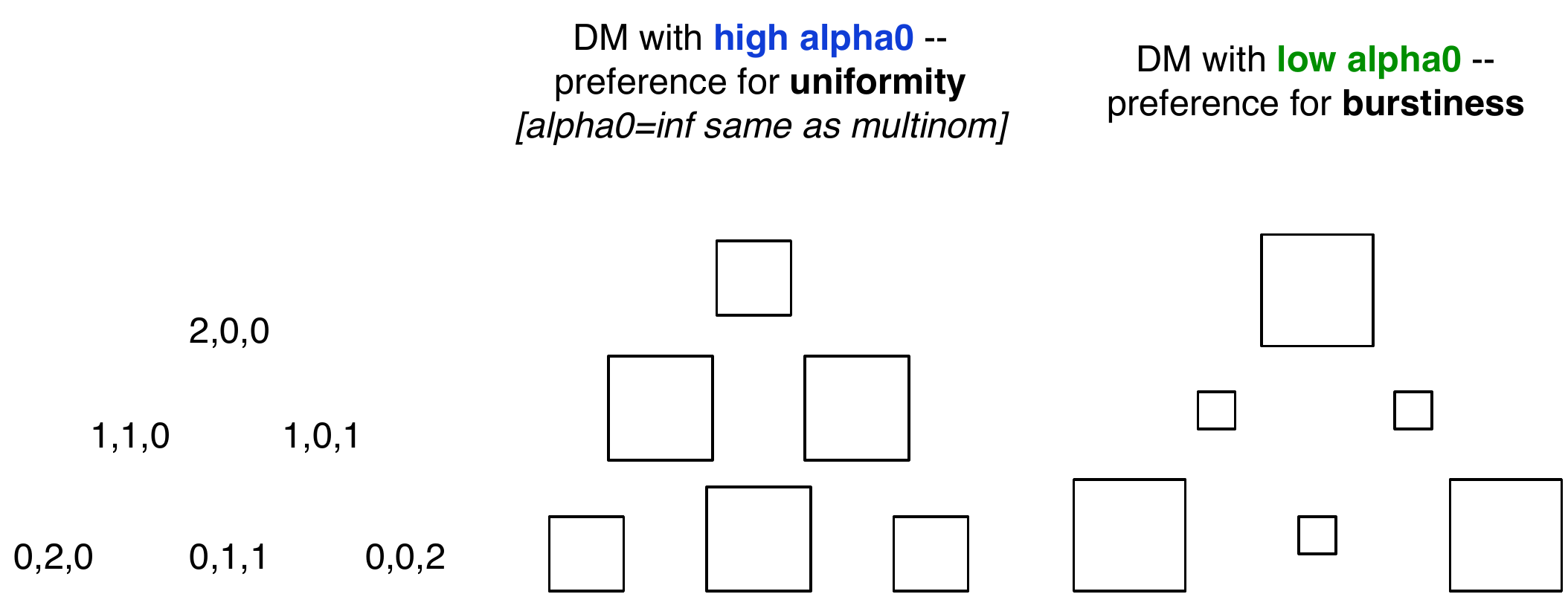}
\caption{For two throws of a three-sided die, the six possible outcomes
are laid out as shown on \textbf{left} (reminiscent of a simplex).
\textbf{Center:} The PMF of a two-draw multinomial, mean parameter 
$p=(\frac{1}{3}, \frac{1}{3}, \frac{1}{3})$.  Area of box corresponds to probability of the outcome.
\textbf{Right:} The PMF of a Dirichlet-multinomial, with the same mean,
but a low concentration $A$.  No multinomial can represent this distributiom.
\label{f:dm_viz}}
\end{figure}

\noindent
Note the DM gives, for small $\alpha$ priors, a bowed-out preference to count
vectors that lie on the extremes of the $N$-simplex---just like the Dirichlet
with small $\alpha$ prefers bowed-out points on the 1-simplex.  This can be
seen as a preference for ``bursty'' behavior.  A multinomial cannot do this:
you only can specify the mean, then the variance is fixed.  In the DM, you
control both mean and variance ($\approx$ inverse concentration), allowing burstiness
a.k.a. over/under-dispersion.
See Figure~\ref{f:dm_viz} for an illustration.

For a single DM draw where $N=1$, instead of representing the draw as a sparse count vector $x$,
we can represent it instead as an integer ID, $z \in \{1..K\}$.
In the PMF, all the combinatorics drop away, leaving:
\[ DM(z; \alpha) = \frac{\alpha_z}{A} \]

\noindent which is simply the Dirichlet mean parameter.
When there's only one draw there's no such thing as burstiness or not.

It is also easy to derive the posterior predictive distribution for a Dirichlet-Multinomial hierarchical model.
We want to know the probability of the next draw $z$ given the previous draws
$z_{-i}$, using our entire posterior beliefs about $\theta$.
Represent $z_{-i}$ as count vector $\vec{n} = (n_1..n_K)$,
i.e. $n_k=\sum_{j \neq i} 1\{z_j=k\}$:

\begin{align}
p(z|z_{-i}, \vec{\alpha}) 
&= \int p(z|\theta)\ p(\theta|z_{-i},\alpha)\ d\theta \\
&= \int Multinom(z; \theta)\ Dir(\theta; \vec{\alpha}+\vec{n})\ d\theta
\end{align}
The second step used Dirichlet-multinomial conjugacy.  Now this is just the 1-draw DM (i.e.~the mean of the conjugately-updated Dirichlet),
\begin{align}
p(z|z_{-i}, \vec{\alpha}) 
&= DM(z; \vec{\alpha}+\vec{n}) \\
&= \frac{\alpha_z + n_z}{A+N}
\end{align}

\subsection{DM PMF in LDA hyperparameter sampling}
\label{hyper_dcm}

Going through the full DM is not necessary to derive the collapsed Gibbs sampling equations,
but it \emph{is} necessary for hyperparameter sampling, which requires evaluating the likelihood of the
entire dataset under different hyperparamters $\alpha$.  LDA under collapsing
can be viewed as a series of DM draws:

\begin{itemize}
% \item For priors $\alpha$, $\beta$
\item For each $d$, sample vector $z_{\{i:\ d_i=d\}} \sim DMPath(\alpha)$
\item For each $k$, sample vector $w_{\{i:\ z_i=k\}} \sim DMPath(\beta)$
\end{itemize}

\noindent where ``DMPath'' indicates choosing one random sequence having
the counts of one DM draw; its PMF is the $DM1$ function of its count vector.
(This could be computed by proceeding through a Polya urn process a.k.a.~(finite) Chinese restaurant process.)
Therefore, for their Gibbs update steps, the hyperparameter likelihoods are:

\begin{align}
% p(\alpha \mid z,w) &\propto 
p(z \mid \alpha)  &
= \prod_d DMPath(z_{\{i:\ d_i=d\}}; \alpha) \\
% p(\beta \mid z,w) &\propto 
p(w \mid z,\beta)  &
= \prod_k DMPath(w_{\{i:\ z_i=k\}}; \beta)
\end{align}

\noindent
For the other models, analogous formulations are available as well.
We were initially tempted
to try to compute the likehoods with per-token local conditionals similar to what is used for the
$z$ Gibbs updates,
\[\prod_i p(w_i|z_i, \beta)\ p(z_i|z_{-i}; \alpha) \]
\noindent
which is easy to compute, but unfortunately wrong:
it is actually a pseudolikelihood approximation to the likelihood (Besag 1975).
Since it is possible to compute the actual likelihood 
closed-form log-gammas, we do so.

\begin{figure}[t]
  \begin{center}
    \includegraphics[width=3in]{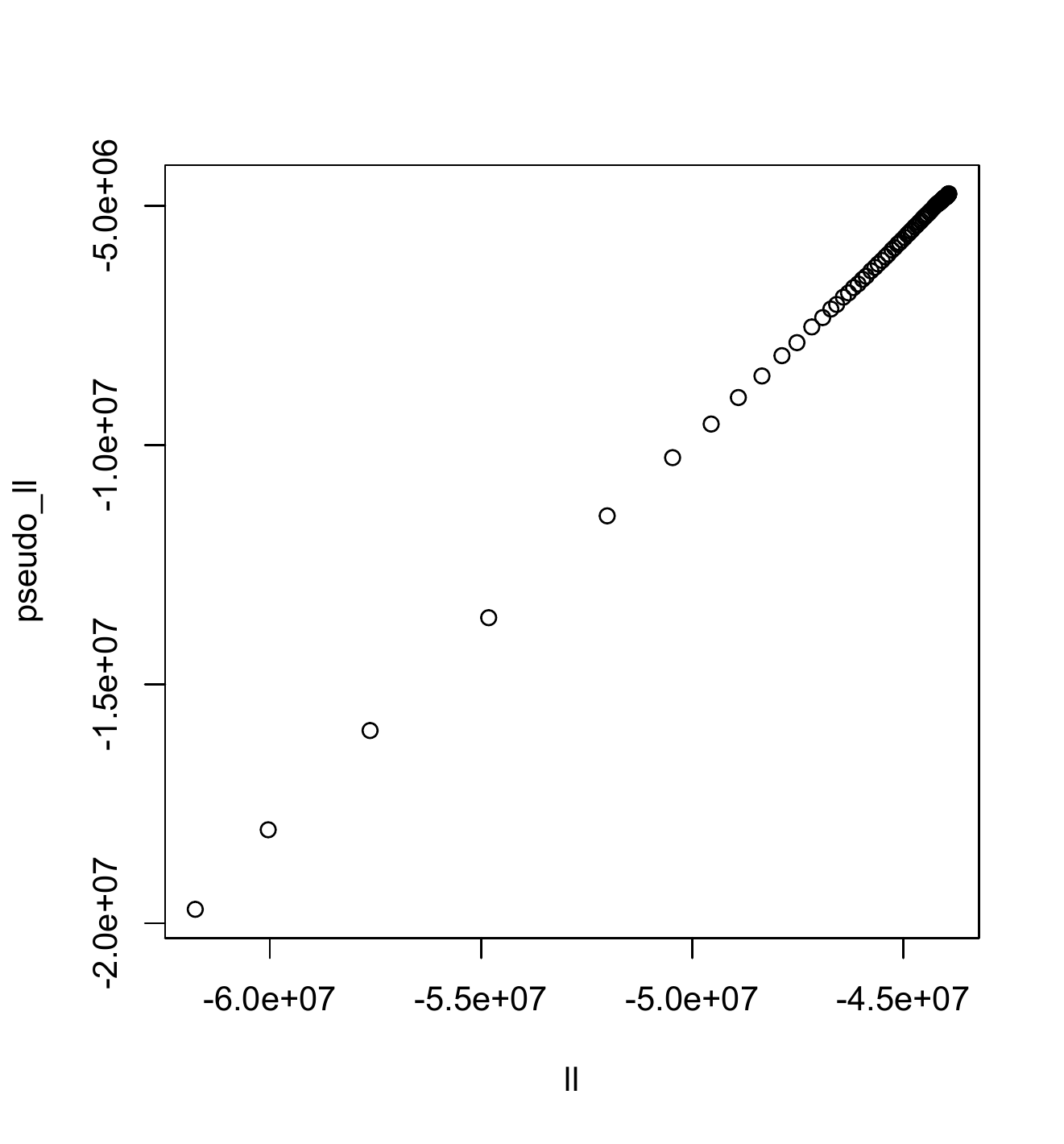}
    \includegraphics[width=3in]{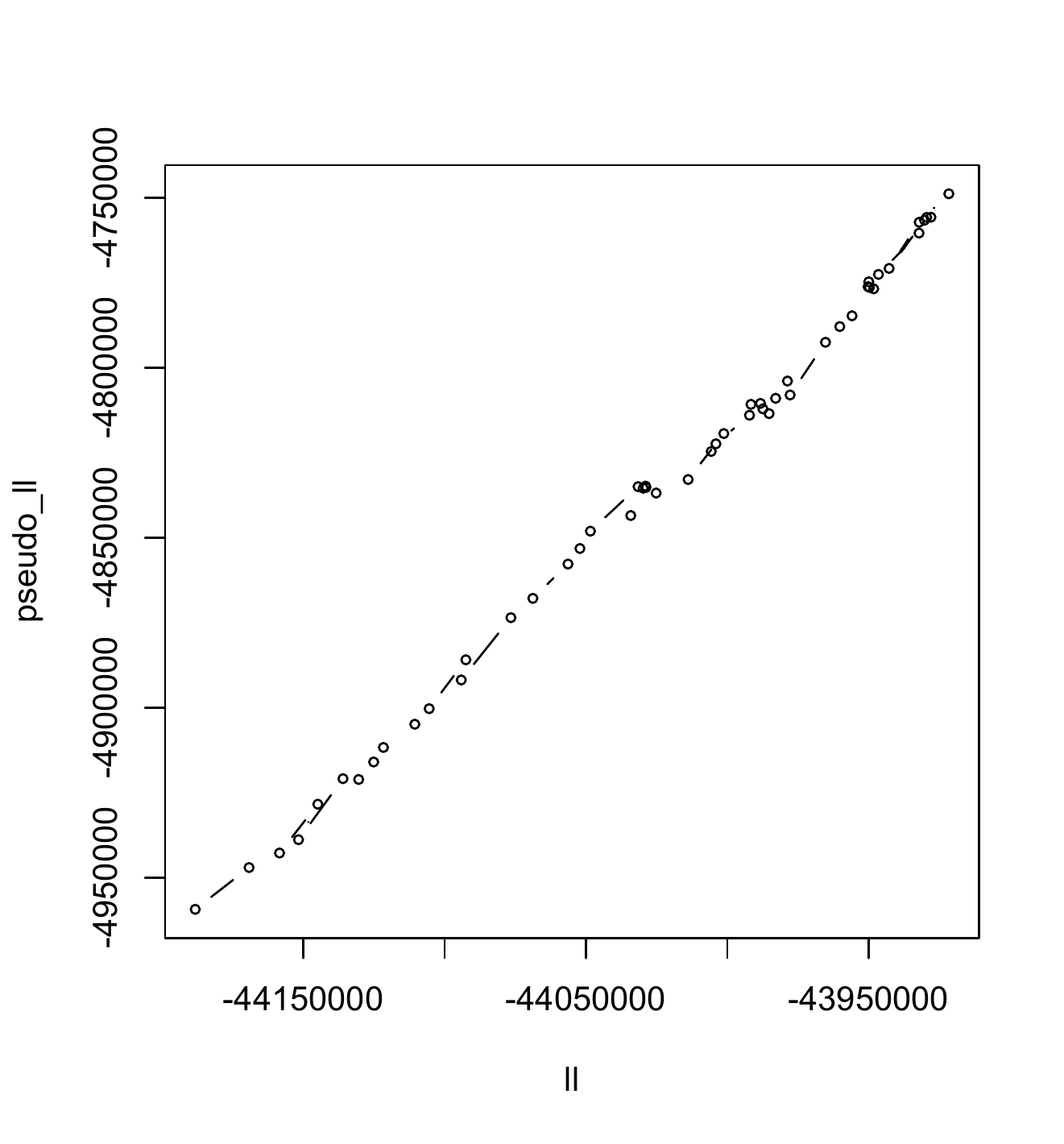}
  \end{center}
  \caption{Correlation of running pseudolikelihood (evaluated during Gibbs sampling) to actual likelihood (evaluated exactly via the DM PMF (section \ref{dcm_math})), for one MCMC run (small dataset, Model 1).
Left: shown for all iterations where likelihood was evaluated.  
Right: shown for iterations 500 and later, with lines drawn between successive iterations.}
  \label{fig:pseudo_corr}
\end{figure}

However, it does turn out the running-sum pseudolikelihood is a good approximation, as shown in Figure \ref{fig:pseudo_corr}, for correlating to likelihood across MCMC samples,
thus could be used to aid MCMC convergence diagnosis
in situations where full likelihood evaluation is computationally expensive
or difficult to implement.

% (We do use sometimes use a running form of this pseudolikelihood
% to diagnose whether the $z$ Gibbs updates are converging or not,
% since it is cheap to compute the resulting $p(w_i,z_i|z_{-i})$
% token likelihood after finishing the $z$ sampling operation.)

\bibliographystyle{plainnat}
\bibliography{brendan-zotero,manual,newbib}

\end{document}